\documentclass{article}

\usepackage[
	activate={true,nocompatibility},
	tracking=true
]{microtype}

\usepackage{amsmath}
\usepackage{amssymb}
\newcommand{\given}{\,|\,}{}

\usepackage{graphicx}
\usepackage{subfig}
\usepackage{booktabs}
\usepackage[group-separator={,}]{siunitx}
\usepackage{listings}

\usepackage{FiraMono}
\usepackage[T1]{fontenc}

\usepackage{enumitem}
\setitemize{noitemsep,topsep=0pt,parsep=0pt,partopsep=0pt}

\usepackage{hyperref}

\usepackage[accepted]{icml2020}

\icmltitlerunning{Zero-Shot Text-to-Image Generation}

\begin{document}

\twocolumn[
\icmltitle{Zero-Shot Text-to-Image Generation}

\begin{icmlauthorlist}
\icmlauthor{Aditya Ramesh}{oai}
\icmlauthor{Mikhail Pavlov}{oai}
\icmlauthor{Gabriel Goh}{oai}
\icmlauthor{Scott Gray}{oai} \\
\icmlauthor{Chelsea Voss}{oai}
\icmlauthor{Alec Radford}{oai}
\icmlauthor{Mark Chen}{oai}
\icmlauthor{Ilya Sutskever}{oai}
\end{icmlauthorlist}

\icmlaffiliation{oai}{OpenAI, San Francisco, California, United States}
\icmlcorrespondingauthor{Aditya Ramesh}{\_@adityaramesh.com}

\icmlkeywords{Machine Learning, ICML}
\vskip 0.3in
]
\printAffiliationsAndNotice{}
\begin{abstract}
Text-to-image generation has traditionally focused on finding better modeling assumptions for training on a fixed dataset. These assumptions might involve complex architectures, auxiliary losses, or side information such as object part labels or segmentation masks supplied during training. We describe a simple approach for this task based on a transformer that autoregressively models the text and image tokens as a single stream of data. With sufficient data and scale, our approach is competitive with previous domain-specific models when evaluated in a zero-shot fashion.
\end{abstract}

\section{Introduction}
\label{intro}

Modern machine learning approaches to text to image synthesis started with the work of \citet{mansimov2015generating}, who showed that the~DRAW \citet{gregor2015draw} generative model, when extended to condition on image captions, could also generate novel visual scenes. \citet{reed2016generative} later demonstrated that using a generative adversarial network \citep{goodfellow2014generative}, rather than a recurrent variational auto-encoder, improved image fidelity. \citet{reed2016generative} showed that this system could not only generate objects with recognizable properties, but also could \textit{zero-shot} generalize to held-out categories.

Over the next few years, progress continued using a combination of methods. These include improving the generative model architecture with modifications like multi-scale generators \citep{zhang2017stackgan,zhang2018stackgan++}, integrating attention and auxiliary losses \citep{xu2018attngan}, and leveraging additional sources of conditioning information beyond just text \citep{reed2016learning,li2019object,koh2021text}.

\begin{figure}
    \centering
    \includegraphics[width=\linewidth]{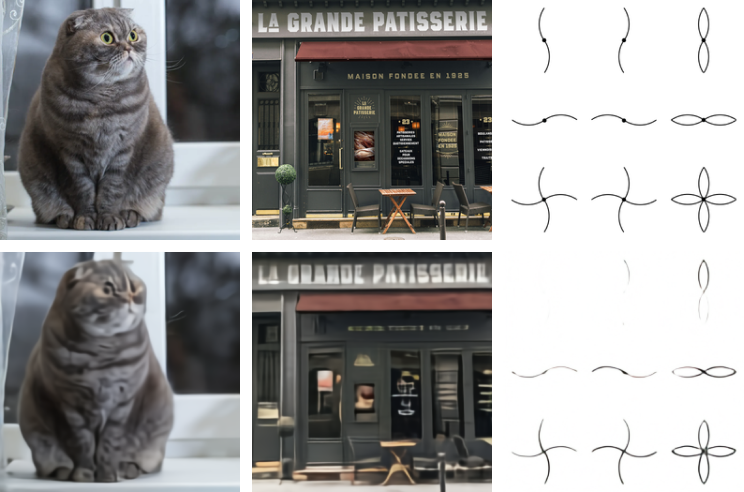}
    \caption{Comparison of original images (top) and reconstructions from the discrete VAE (bottom). The encoder downsamples the spatial resolution by a factor of~8. While details (e.g., the texture of the cat's fur, the writing on the storefront, and the thin lines in the illustration) are sometimes lost or distorted, the main features of the image are still typically recognizable. We use a large vocabulary size of~8192 to mitigate the loss of information.}
    \label{fig:dvae_rec}
\end{figure}
Separately, \citet{nguyen2017plug} propose an energy-based framework for conditional image generation that obtained a large improvement in sample quality relative to contemporary methods. Their approach can incorporate pretrained discriminative models, and they show that it is capable of performing text-to-image generation when applied to a captioning model pretrained on MS-COCO.
More recently, \citet{cho2020x} also propose a method that involves optimizing the input to a pretrained cross-modal masked language model. While significant increases in visual fidelity have occurred as a result of the work since \citet{mansimov2015generating}, samples can still suffer from severe artifacts such as object distortion, illogical object placement, or unnatural blending of foreground and background elements.

Recent advances fueled by large-scale generative models suggest a possible route for further improvements. Specifically, when compute, model size, and data are scaled carefully, autoregressive transformers \citep{vaswani2017attention} have achieved impressive results in several domains such as text \citep{radford2019language}, images \citep{chen2020generative}, and audio \citep{dhariwal2020jukebox}.

\begin{figure*}
\centering
\captionsetup[subfigure]{width=1.5in}
\subfloat[a tapir made of accordion. a tapir with the texture of an accordion.]{
  \setlength\extrarowheight{-3pt}
\begin{tabular}{@{}c@{\hskip 0.05in} c@{}}
  \fbox{\includegraphics[scale=0.2]{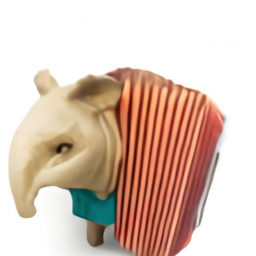}}
  & \fbox{\includegraphics[scale=0.2]{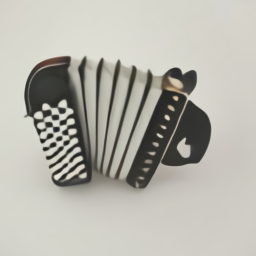}} 
 \\
  \fbox{\includegraphics[scale=0.2]{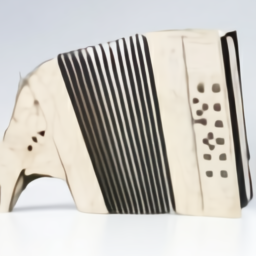}} 
 & \fbox{\includegraphics[scale=0.2]{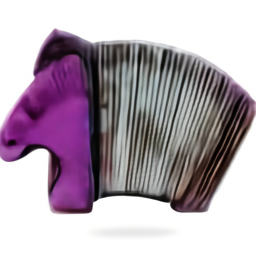}} 
 \\
\end{tabular}\label{fig:blog_samples1}}%
\hspace{0.75mm}%
\subfloat[an illustration of a baby hedgehog in a christmas sweater walking a dog]{  \setlength\extrarowheight{-3pt}
\begin{tabular}{@{}c@{\hskip 0.05in}c@{}}
  \fbox{\includegraphics[scale=0.2]{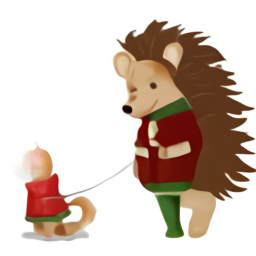}}
  & \fbox{\includegraphics[scale=0.2]{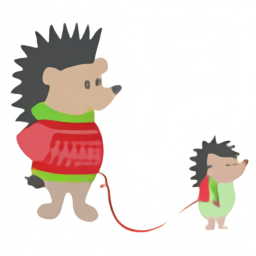}} 
 \\
  \fbox{\includegraphics[scale=0.2]{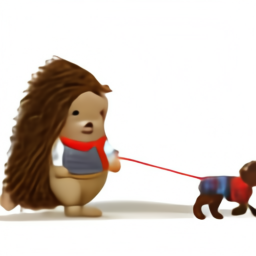}} 
 & \fbox{\includegraphics[scale=0.2]{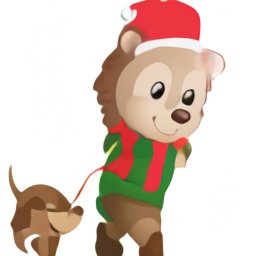}} 
 \\
\end{tabular}\label{fig:blog_samples2}}%
\hspace{0.75mm}%
\subfloat[a neon sign that reads ``backprop''. a neon sign that reads ``backprop''. backprop neon sign]{
  \setlength\extrarowheight{-3pt}
\begin{tabular}{@{}c@{\hskip 0.05in} c@{}}
  \fbox{\includegraphics[scale=0.2]{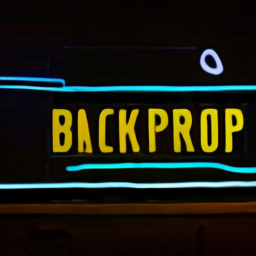}}
  & \fbox{\includegraphics[scale=0.2]{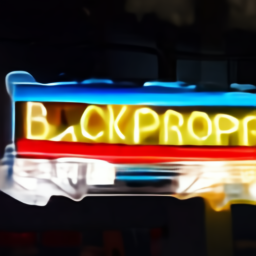}} 
 \\
  \fbox{\includegraphics[scale=0.2]{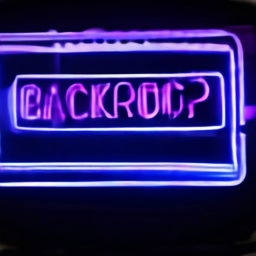}} 
 & \fbox{\includegraphics[scale=0.2]{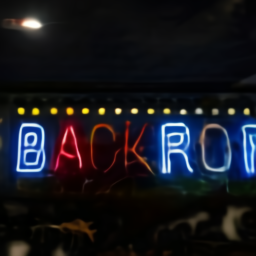}} 
 \\
\end{tabular}\label{fig:blog_samples3}}%
\hspace{0.75mm}%
\subfloat[the exact same cat on the top as a sketch on the bottom]{
  \setlength\extrarowheight{-3pt}
\begin{tabular}{@{}c@{\hskip 0.05in}c@{}}
  \fbox{\includegraphics[scale=0.2]{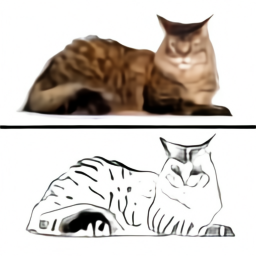}}
  & \fbox{\includegraphics[scale=0.2]{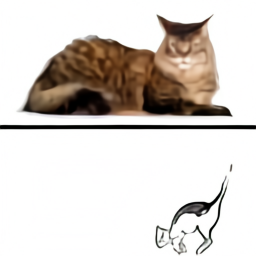}} 
 \\
  \fbox{\includegraphics[scale=0.2]{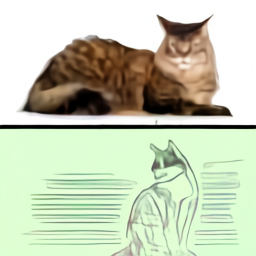}} 
 & \fbox{\includegraphics[scale=0.2]{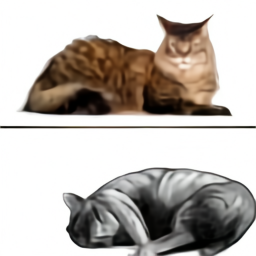}} 
 \\
\end{tabular}\label{fig:blog_samples4}}%
\caption{With varying degrees of reliability, our model appears to be able to combine distinct concepts in plausible ways, create anthropomorphized versions of animals, render text, and perform some types of image-to-image translation.}
\label{fig:blog_samples}
\end{figure*}
%
By comparison, text-to-image generation has typically been evaluated on relatively small datasets such as MS-COCO and CUB-200 \citep{welinder2010caltech}. Could dataset size and model size be the limiting factor of current approaches? In this work, we demonstrate that training a 12-billion parameter autoregressive transformer on 250~million image-text pairs collected from the internet results in a flexible, high fidelity generative model of images controllable through natural language.

The resulting system achieves high quality image generation on the popular MS-COCO dataset \textit{zero-shot}, without using any of the training labels. It is preferred over prior work trained on the dataset by human evaluators 90\% of the time. We also find that it is able to perform complex tasks such as image-to-image translation at a rudimentary level. This previously required custom approaches \citep{isola2017image}, rather
emerging as a capability of a single, large generative model.

\begin{figure*}[t]
    \centering
    \includegraphics[width=\linewidth]{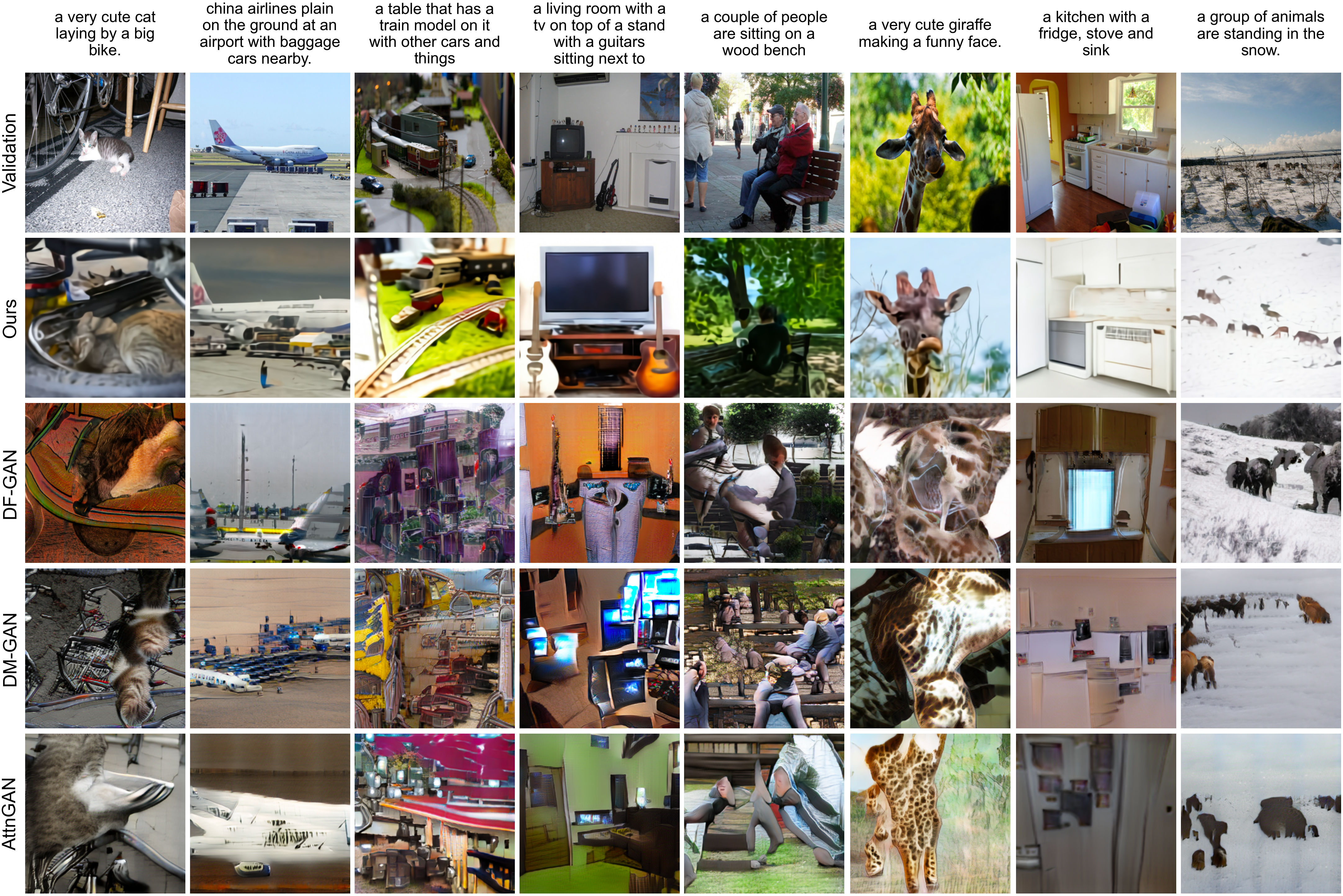}
    \caption{Comparison of samples from our model to those from prior approaches on captions from MS-COCO. Each of our model samples is the best of~512 as ranked by the contrastive model. We do not use any manual cherrypicking with the selection of either the captions or the samples from any of the models.}
    \label{fig:coco_cmp}
\end{figure*}
\section{Method}
Our goal is to train a transformer~\cite{vaswani2017attention} to autoregressively model the text and image tokens as a single stream of data. However, using pixels directly as image tokens would require an inordinate amount of memory for high-resolution images. Likelihood objectives tend to prioritize modeling short-range dependencies between pixels \citep{salimans2017pixelcnn++}, so much of the modeling capacity would be spent capturing high-frequency details instead of the low-frequency structure that makes objects visually recognizable to us.

We address these issues by using a two-stage training procedure, similar to~\cite{oord2017neural,razavi2019generating}:
\begin{itemize}
    \item \textbf{Stage~1.} We train a discrete variational autoencoder~(dVAE)\footnote{\url{https://github.com/openai/DALL-E}} to compress each $256 \times 256$ RGB image into a $32 \times 32$ grid of image tokens, each element of which can assume \num{8192} possible values. This reduces the context size of the transformer by a factor of~$192$ without a large degradation in visual quality (see Figure~\ref{fig:dvae_rec}).
    \item \textbf{Stage~2.} We concatenate up to~256 BPE-encoded text tokens with the~$32 \times 32 = 1024$ image tokens, and train an autoregressive transformer to model the joint distribution over the text and image tokens.
\end{itemize}

The overall procedure can be viewed as maximizing the evidence lower bound (ELB)~\cite{kingma2013auto,rezende2014stochastic} on the joint likelihood of the model distribution over images~$x$, captions~$y$, and the tokens~$z$ for the encoded RGB image. We model this distribution using the factorization $p_{\theta, \psi}(x, y, z) = p_\theta(x \given y, z) p_\psi(y, z)$, which yields the lower bound
\begin{multline}
    \ln p_{\theta, \psi}(x, y) \geqslant \!\!\!\!\!\!\!\! \mathop{\mathbb{E}}_{\substack{\vspace{0.1mm} \\ z \sim q_\phi(z \given x)}} \!\!\!\!\!\!\!\! \big( \ln p_\theta(x \given y, z)
    \;- \\ 
    \beta\, D_{\mathrm{KL}}(q_\phi(y, z \given x), p_\psi(y,z)) \big), \label{eq:elb}
\end{multline}
where:
\begin{itemize}
    \item $q_\phi$ denotes the distribution over the~$32 \times 32$ image tokens generated by the dVAE encoder given the RGB image~$x$\footnote{We assume that $y$ is conditionally independent of~$x$ given~$z$.};
    \item $p_\theta$ denotes the distribution over the RGB images generated by the dVAE decoder given the image tokens; and
    \item $p_\psi$ denotes the joint distribution over the text and image tokens modeled by the transformer.
\end{itemize}
Note that the bound only holds for $\beta=1$, while in practice we find it helpful to use larger values~\cite{higgins2016beta}. The following subsections describe both stages in further detail.\footnote{In preliminary experiments on ImageNet~\cite{deng2009imagenet}, we attempted to maximize the ELB with respect to~$\phi$, $\theta$, and~$\psi$ jointly, but were unable to improve on two-stage training.}

%
\subsection{Stage One: Learning the Visual Codebook}

In the first stage of training, we maximize the ELB with respect to~$\phi$ and~$\theta$, which corresponds to training a~dVAE on the images alone. We set the initial prior~$p_\psi$ to the uniform categorical distribution over the~$K = \num{8192}$ codebook vectors, and $q_\phi$ to be categorical distributions parameterized by the~\num{8192} logits at the same spatial position in the~$32 \times 32$ grid output by the encoder.

The ELB now becomes difficult to optimize: as $q_\psi$ is a discrete distribution, and we cannot use the reparameterization gradient to maximize it. \citet{oord2017neural,razavi2019generating} address this using an online cluster assignment procedure coupled with the straight-through estimator~\cite{bengio2013estimating}. We instead use the gumbel-softmax relaxation~\cite{jang2016categorical,maddison2016concrete}, replacing the expectation over $q_\phi$ with one over $q^\tau_\phi$, where the relaxation becomes tight as the temperature~$\tau \to 0$. The likelihood for $p_\theta$ is evaluated using the log-laplace distribution (see Appendix~\ref{sec:dvae_logit_laplace} for a derivation).




The relaxed ELB is maximized using Adam~\cite{kingma2014adam} with exponentially weighted iterate averaging. Appendix~\ref{sec:dvae_train} gives a complete description of the hyperparameters, but we found the following to be especially important for stable training:
\begin{itemize}
    \item Specific annealing schedules for the relaxation temperature and step size. We found that annealing $\tau$ to~$1 / 16$ was sufficient to close the gap between the relaxed validation ELB and the true validation ELB with~$q_\phi$ intsead of~$q_\phi^\tau$.
    \item The use of $1 \times 1$ convolutions at the end of the encoder and the beginning of the decoder. We found that reducing the receptive field size for the convolutions around the relaxation led to it generalizing better to the true ELB.
    \item Multiplication of the outgoing activations from the encoder and decoder resblocks by a small constant, to ensure stable training at initialization.
\end{itemize}
We also found that increasing the KL weight to~$\beta = 6.6$ promotes better codebook usage and ultimately leads to a \emph{smaller} reconstruction error at the end of training.\footnote{This is contrary to the usual tradeoff between the two terms. We speculate that for smaller values of~$\beta$, the noise from the relaxation causes the optimizer to reduce codebook usage toward the beginning of training, resulting in worse ELB at convergence.}
\subsection{Stage Two: Learning the Prior}
\label{sec:learning_prior}
In the second stage, we fix~$\phi$ and~$\theta$, and learn the prior distribution over the text and image tokens by maximizing the ELB with respect to~$\psi$. Here, $p_\psi$ is represented by a 12-billion parameter sparse transformer~\cite{child2019generating}.

Given a text-image pair, we BPE-encode~\cite{sennrich2015neural} the lowercased caption using at most 256 tokens\footnote{During training, we apply~10\% BPE dropout~\cite{provilkov2019bpe}, whose use is common in the neural machine translation literature.} with vocabulary size \num{16384}, and encode the image using $32 \times 32 = 1024$ tokens with vocabulary size \num{8192}. The image tokens are obtained using argmax sampling from the dVAE encoder logits, without adding any gumbel noise.\footnote{Strictly speaking, Equation~\ref{eq:elb} requires us to sample from the categorical distribution specified by the dVAE encoder logits, rather than taking the argmax. In preliminary experiments on ImageNet, we found that this was a useful regularizer in the overparameterized regime, and allows the transformer to be trained using soft targets for the cross-entropy loss. We decided against this here since the model in consideration is in the underparameterized regime.} Finally, the text and image tokens are concatenated and modeled autoregressively as a single stream of data.

The transformer is a decoder-only model in which each image token can attend to all text tokens in any one of its 64 self-attention layers. The full architecture is described in Appendix~\ref{sec:xf_arch}. There are three different kinds of self-attention masks used in the model. The part of the attention masks corresponding to the text-to-text attention is the standard causal mask, and the part for the image-to-image attention uses either a row, column, or convolutional attention mask.\footnote{We found using a single attention operation for all three interactions -- ``text attends to text'', ``image attends to text'', and ``image attends to image'' -- to perform better than using separate attention operations that are independently normalized.} 

We limit the length of a text caption to 256~tokens, though it is not totally clear what to do for the ``padding'' positions in between the last text token and the start-of-image token. One option is to set the logits for these tokens to~$-\infty$ in the self-attention operations. Instead, we opt to learn a special padding token separately for each of the~256 text positions. This token is used only when no text token is available. In preliminary experiments on Conceptual Captions~\cite{sharma2018conceptual}, we found that this resulted in higher validation loss, but better performance on out-of-distribution captions.

We normalize the cross-entropy losses for the text and image tokens by the total number of each kind in a batch of data. Since we are primarily interested in image modeling, we multiply the cross-entropy loss for the text by~$1 / 8$ and the cross-entropy loss for the image by~$7 / 8$. The objective is optimized using Adam with exponentially weighted iterate averaging; Appendix~\ref{sec:xf_train} describes the training procedure in more detail. We reserved about~\num{606000} images for validation, and found no signs of overfitting at convergence.

\begin{figure}[t]
    \centering
    \includegraphics[width=\linewidth]{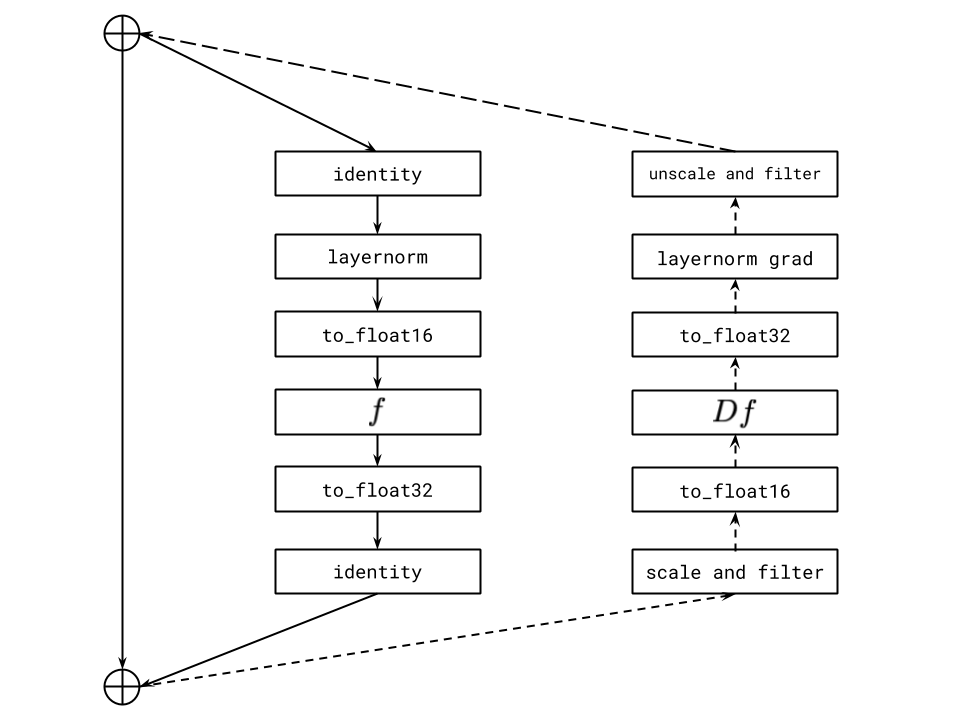}
    \caption{Illustration of per-resblock gradient scaling for a transformer resblock. The solid line indicates the sequence of operations for forward propagation, and the dashed line the sequence of operations for backpropagation. We scale the incoming gradient for each resblock by its gradient scale, and unscale the outgoing gradient before it is added to the sum of the gradients from the successive resblocks. The activations and gradients along the identity path are stored in 32-bit precision. The ``filter'' operation sets all Inf and NaN values in the activation gradient to zero. Without this, a nonfinite event in the current resblock would cause the gradient scales for all preceding resblocks to unnecessarily drop, thereby resulting in underflow.}
    \label{fig:grad_scaling}
\end{figure}

\begin{figure}[t]
    \centering
    \includegraphics[width=3in]{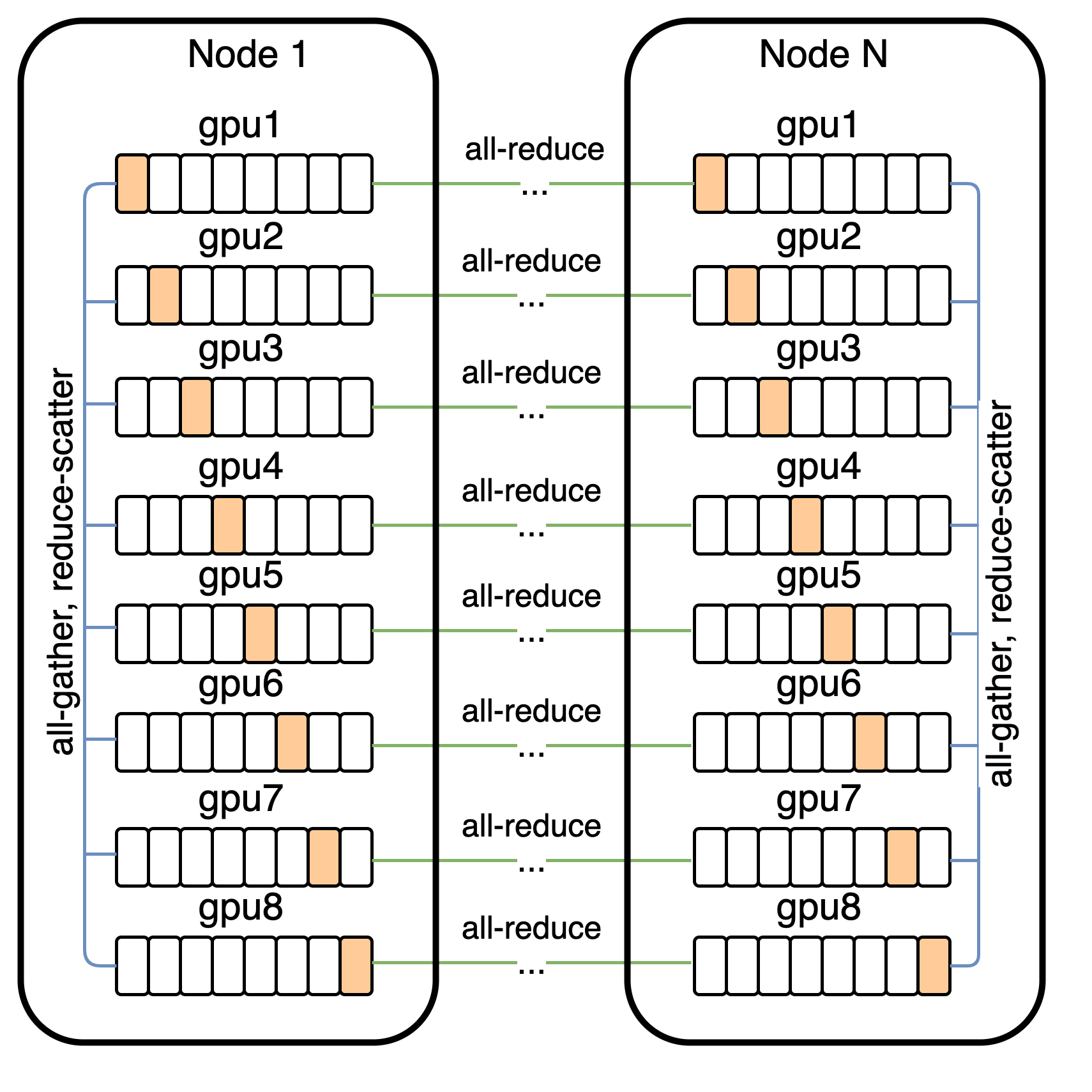}
    \caption{Communication patterns used for distributed training. Each parameter array in the model is sharded among the eight GPUs on each machine. During forward propagation, we prefetch the parameter shards for the next resblock (using all-gather) while computing the activations for the current resblock. To conserve memory, the parameter shards from the other GPUs are immediately discarded. Similarly, during backpropagation, we prefetch the parameter shards for the previous resblock while computing the activations and gradients for the current resblock. After all GPUs have computed the gradient with respect to an all-gathered parameter, the reduce-scatter operation leaves each GPU with only one slice -- i.e., the gradient for its parameter shard, averaged over the eight GPUs.}
    \label{fig:dist_comm}
\end{figure}

\subsection{Data Collection}
\label{sec:data_collection}

Our preliminary experiments for models up to $1.2$~billion parameters were carried out on Conceptual Captions, a dataset of 3.3~million text-image pairs that was developed as an extension to MS-COCO~\cite{lin2014microsoft}. 

To scale up to $12$-billion parameters, we created a dataset of a similar scale to JFT-300M~\cite{sun2017revisiting} by collecting 250~million text-images pairs from the internet. This dataset does not include MS-COCO, but does include Conceptual Captions and a filtered subset of YFCC100M~\cite{thomee2016yfcc100m}. As MS-COCO was created from the latter, our training data includes a fraction of the MS-COCO validation images (but none of the captions). We control for this in the quantitative results presented in Section~\ref{sec:experiments} and find that it has no appreciable bearing on the results. We provide further details about the data collection process in Appendix~\ref{sec:data_collection_details}.

\subsection{Mixed-Precision Training}
\label{sec:mp_train}

To save GPU~memory and increase throughput, most parameters, Adam moments, and activations are stored in 16-bit precision. We also use activation checkpointing and recompute the activations within the resblocks during the backward pass. Getting the model to train in 16-bit precision past one billion parameters, without diverging, was the most challenging part of this project.

We believe the root cause of this instability to be underflow in the 16-bit gradients. Appendix~\ref{sec:mp_train_guidelines} presents a set of guidelines we developed to avoid underflow when training large-scale generative models. Here, we describe one of these guidelines: per-resblock gradient scaling.

Similar to prior work~\cite{liu2020understanding}, we found that the norms of the activation gradients from the resblocks decrease monotonically as we move from the earlier resblocks to the later ones.\footnote{It is possible that better initialization schemes~\cite{liu2020understanding} might be able to avoid this, but we did not have success with alternative schemes in our experiments.} As the model is made deeper and wider, the true exponents of the activation gradients for later resblocks can fall below the minimum exponent of the 16-bit format. Consequently, they get rounded to zero, a phenomenon called \emph{underflow}. We found that eliminating underflow allowed for stable training to convergence.

Standard loss scaling~\cite{micikevicius2017mixed} is able to avoid underflow when the range spanned by the smallest and largest activation gradients (in absolute value) fits within the exponent range of the 16-bit format. On NVIDIA V100 GPUs, this exponent range is specified by five bits. While this is sufficient for training vanilla language models of the same size, we found the range to be too small for the text-to-image model.

Our fix, which is shown in Figure~\ref{fig:grad_scaling}, involves using a separate ``gradient scale'' for each resblock in the model. This can be seen as a practical alternative to a more general framework for mixed-precision training called Flexpoint~\cite{koster2017flexpoint}, with the advantage that specialized GPU kernels are not required. We found that \citet{sun2020ultra} had independently developed similar procedure for training convolutional networks in 4-bit precision.

\subsection{Distributed Optimization}
\label{sec:dist_opt}
\begin{table}[]
    \centering\scriptsize
    \begin{tabular}{ccc}
        \toprule
         Effective Parameter Count & Compression Rank & Compression Rate \\
         \midrule
         $2.8 \cdot 10^9$ ($d_{\mathrm{model}} = 1920$) & 512 & $\approx\! 83\%$ \\
         $5.6 \cdot 10^9$ ($d_{\mathrm{model}} = 2688$) & 640 & $\approx\! 85\%$ \\
         $12.0 \cdot 10^9$ ($d_{\mathrm{model}} = 3968$) & 896 & $\approx\! 86\%$ \\
         \bottomrule
    \end{tabular}
    \caption{We show the relationship between model size and the minimum compression rank for the gradients (up to a multiple of 128) necessary to avoid a gap in the training loss during the first~$10\%$ of training. These results suggest that in our setting, we can achieve a compression rate of about~$85\%$, independent of model size.}
    \label{tab:cmp_rank}
\end{table}
\begin{figure*}[t]
    \centering
    \includegraphics[width=\linewidth]{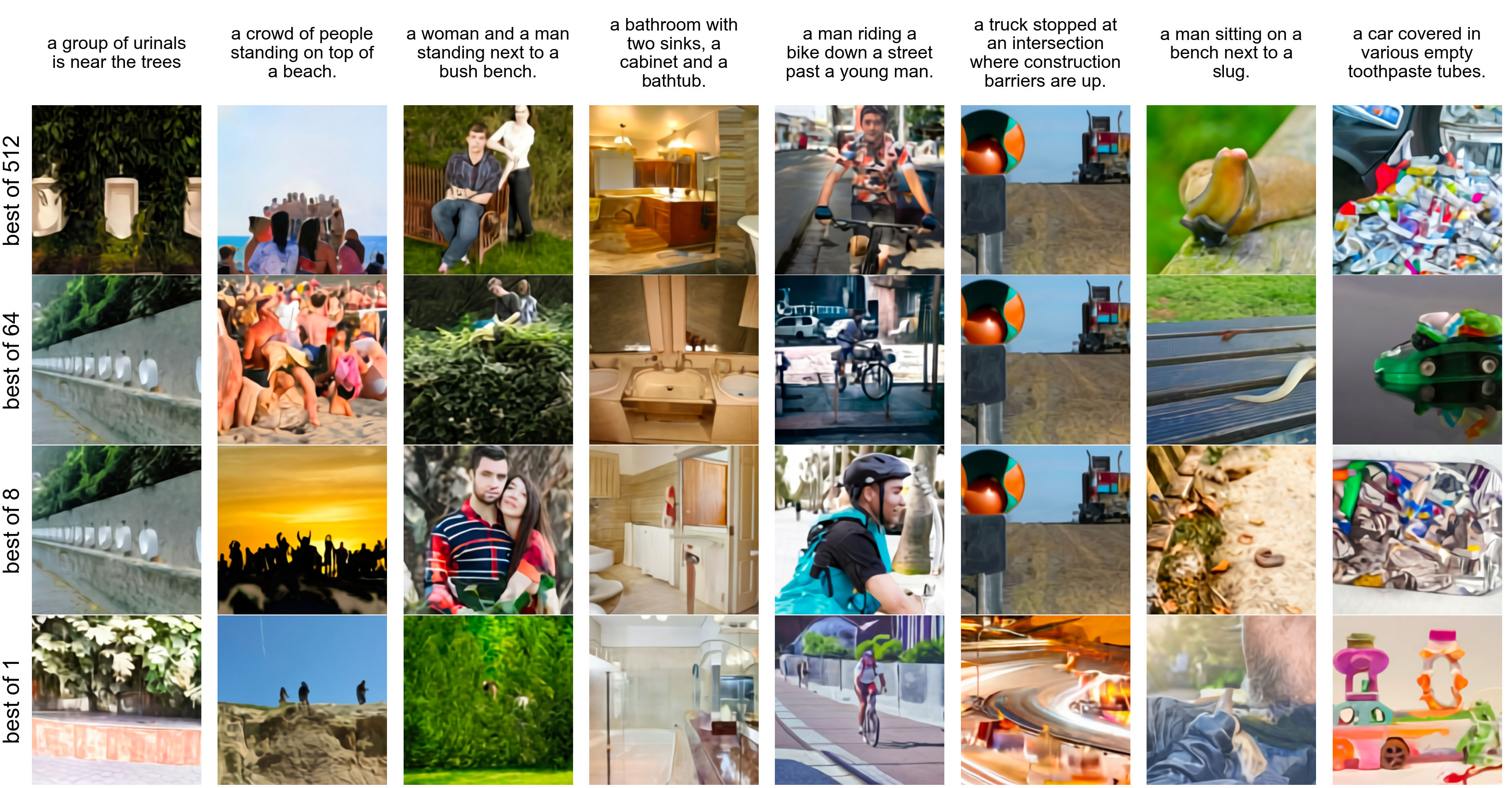}
    \caption{Effect of increasing the number of images for the contrastive reranking procedure on MS-COCO captions.}
    \label{fig:coco_reranking}
\end{figure*}
Our 12-billion parameter model consumes about 24~GB of memory when stored in 16-bit precision, which exceeds the memory of a 16~GB NVIDIA V100 GPU. We address this using parameter sharding~\cite{rajbhandari2019zero}. As shown in Figure~\ref{fig:dist_comm}, parameter sharding allows us to almost completely hide the latency of the intra-machine communication by overlapping it with compute-intensive operations.

On the cluster used to train the model, the bandwidth between machines is much lower than the bandwidth among GPUs on the same machine. This makes the cost of the operation used to average the gradient among the machines (all-reduce) the main bottleneck during training. We were able to drastically reduce this cost by compressing the gradients using PowerSGD~\cite{vogels2019powersgd}.

In our implementation, each GPU in a machine computes the low-rank factors for its parameter shard gradients independently of its neighboring GPUs.\footnote{There is still intra-machine communication for other operations; what we mean is that the low-rank factors across the shards, when concatenated, are not regarded as collectively approximating the gradient for the full parameter matrix.} Once the low-rank factors are computed, each machine sets its error buffer to the residual between the uncompressed gradient averaged over its eight GPUs (obtained from reduce-scatter), and the decompressed gradient obtained from the low-rank factors.

PowerSGD replaces the large communication operation for an uncompressed parameter gradient with two, much smaller communication operations for its low-rank factors. For a given compression rank~$r$ and transformer activation size~$d_\mathrm{model}$, the compression rate is given by~$1 - 5 r / (8 d_\textrm{model})$ (see Appendix~\ref{sec:dist_train_bw}). Table~\ref{tab:cmp_rank} shows that we can achieve a compression rate of about~$85\%$, independent of model size.

In Appendix~\ref{sec:dist_train_impl}, we describe various details that were necessary to get PowerSGD to perform well at scale. These include:
\begin{itemize}
    \item Saving memory by accumulating the gradient into the error buffers during backpropagation, rather than allocating separate buffers.
    \item Minimizing instances in which we zero out the error buffers (e.g., due to nonfinite values encountered during mixed-precision backpropagation, or when resuming training from a checkpoint).
    \item Improving numerical stability by using Householder orthogonalization instead of Gram-Schmidt, together with the addition of a small multiple of the identity matrix to the input.
    \item Avoiding underflow by using a custom 16-bit floating point format for the error buffers, their low-rank factors, and the all-reduce communication operations involving them.
\end{itemize}
We also found the warm-start procedure for the~$Q$ matrix described in \citet{vogels2019powersgd} to be unnecessary: we were able to get equivalent results by fixing~$Q$ to a random gaussian matrix at the start of training, and never updating it.\footnote{We verified that the error in reconstructing the true gradient is higher when~$Q$ is fixed as opposed to being updated using warm-starting, so it is interesting that this does not affect the loss. By contrast, resampling $Q$ at every update causes a large performance hit.}
\subsection{Sample Generation}
Similar to~\citet{razavi2019generating}, we rerank the samples drawn from the transformer using a pretrained contrastive model~\cite{radford2021learning}. Given a caption and a candidate image, the contrastive model assigns a score  based on how well the image matches the caption. Figure~\ref{fig:coco_reranking} shows the effect of increasing the number of samples~$N$ from which we select the top~$k$ images. This process can be seen as a kind of language-guided search~\cite{andreas2017learning}, and is also similar to the auxiliary text-image matching loss proposed by~\citet{xu2018attngan}. Unless otherwise stated, all samples used for both qualitative and quantitative results are obtained without temperature reduction~(i.e., using~$t=1$) (except for Figure~\ref{fig:blog_samples}) and use reranking with~$N = 512$.

\section{Experiments}
\label{sec:experiments}
\begin{figure}[t]
    \centering
    \includegraphics[width=\linewidth]{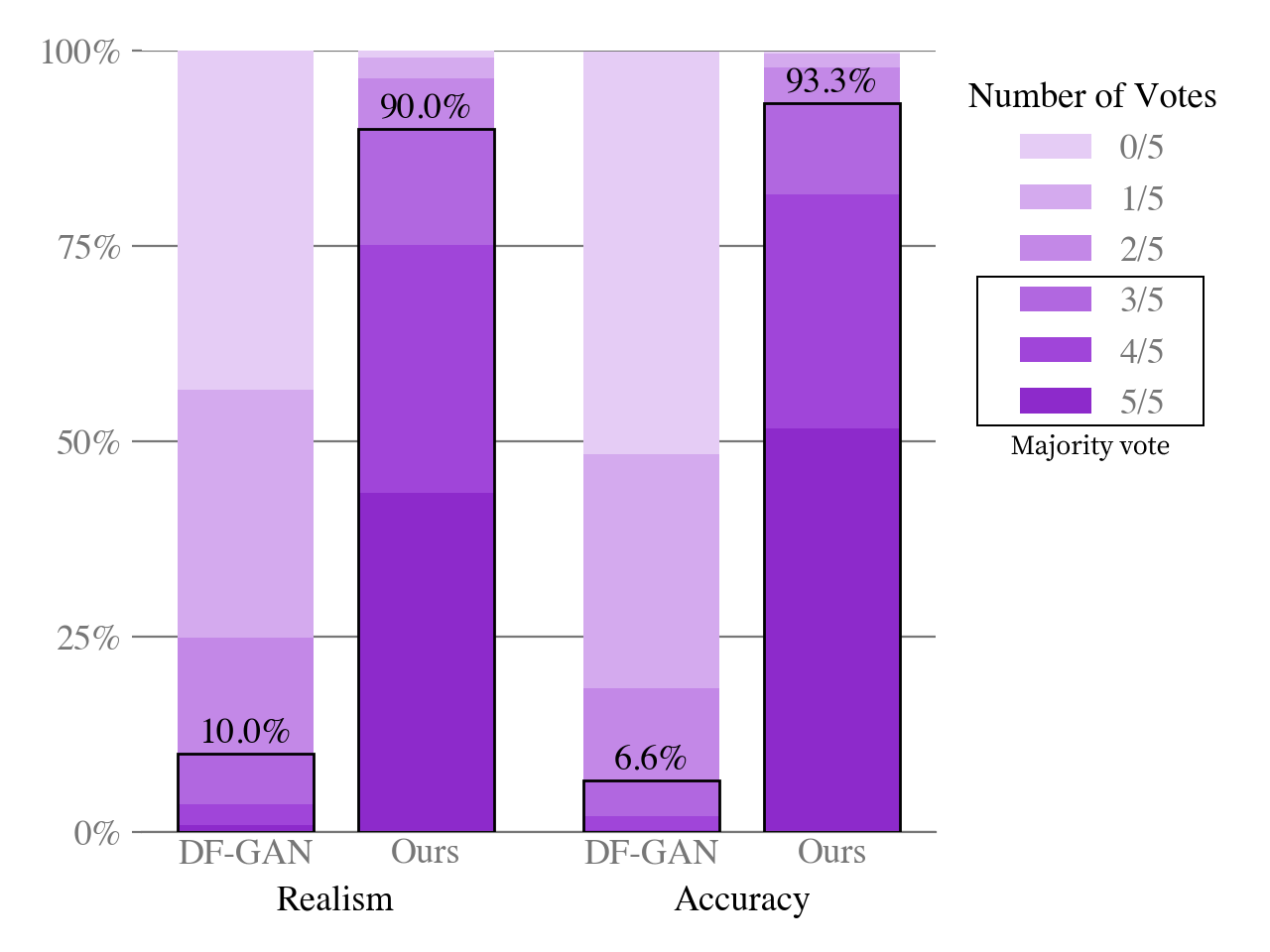}
    \caption{Human evaluation of our model (evaluated zero-shot without temperature reduction) vs prior work~(DF-GAN) on captions from MS-COCO. In a best-of-five vote, our model's sample was chosen as the most realistic 90.0\% of the time, and was chosen as the image best matching a shared caption 93.3\% of the time.}
    \label{fig:human_eval}
\end{figure}
%

\subsection{Quantitative Results}

We evaluate our model zero-shot by comparing it to three prior approaches: AttnGAN~\cite{xu2018attngan}, DM-GAN~\cite{zhu2019dm}, and DF-GAN~\cite{tao2020df}, the last of which reports the best Inception Score~\cite{salimans2016improved} and Fr\'echet Inception Distance~\cite{heusel2017gans} on MS-COCO. Figure~\ref{fig:coco_cmp} qualitatively compares samples from our model to those from prior work.

We also conduct a human evaluation similar to the one used in~\citet{koh2021text} to compare our approach to DF-GAN, the results of which are shown in Figure~\ref{fig:human_eval}. Given a caption, the sample from our model receives the majority vote for better matching the caption~93\% of the time. It also receives the majority vote for being more realistic~90\% of the time.

\begin{figure*}[t]
    \centering
    \captionsetup[subfigure]{width=2in}
    \subfloat[FID and IS on MS-COCO as a function of blur radius.]{%
        \begin{tabular}{@{}c}%
        \includegraphics[width=0.31\linewidth]{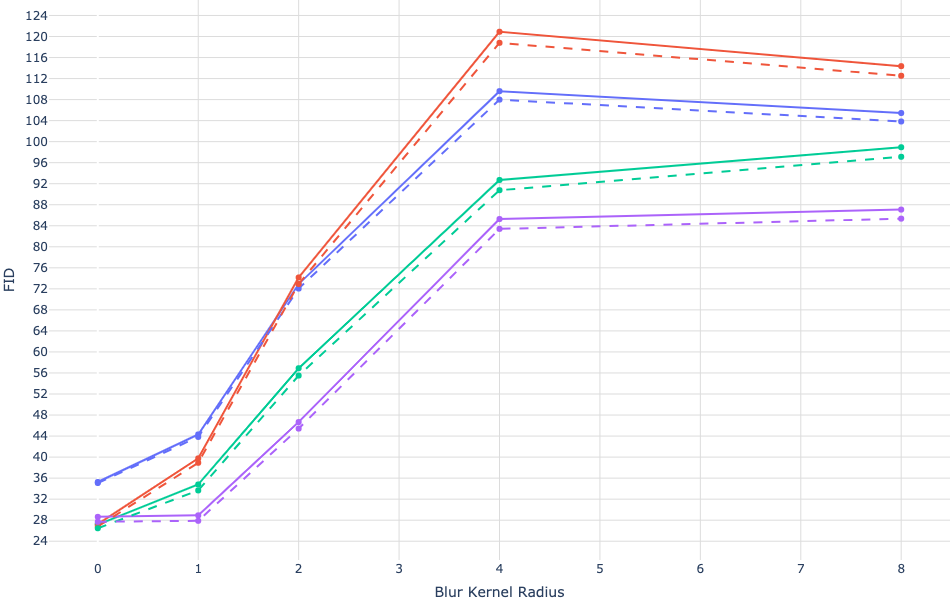} \\
        \includegraphics[width=0.31\linewidth]{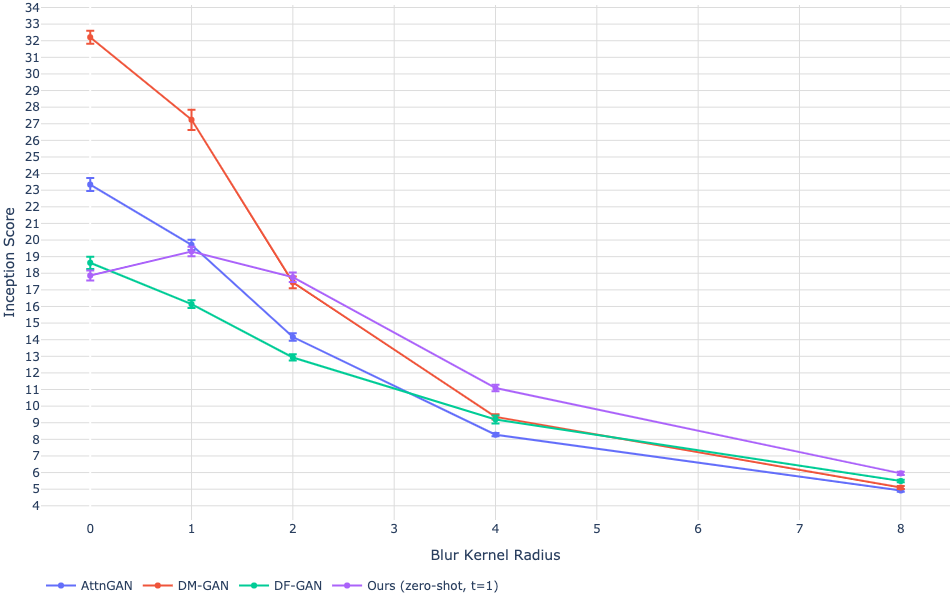}%
        \end{tabular}%
    }\hspace{2mm}
    \subfloat[FID and IS on CUB as a function of blur radius.]{%
        \begin{tabular}{@{}c}%
        \includegraphics[width=0.31\linewidth]{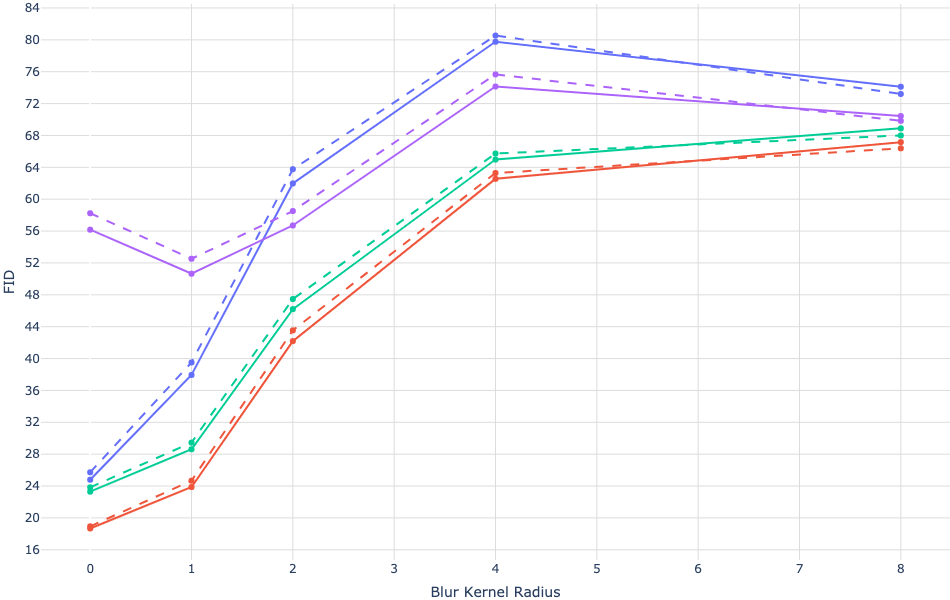} \\
        \includegraphics[width=0.31\linewidth]{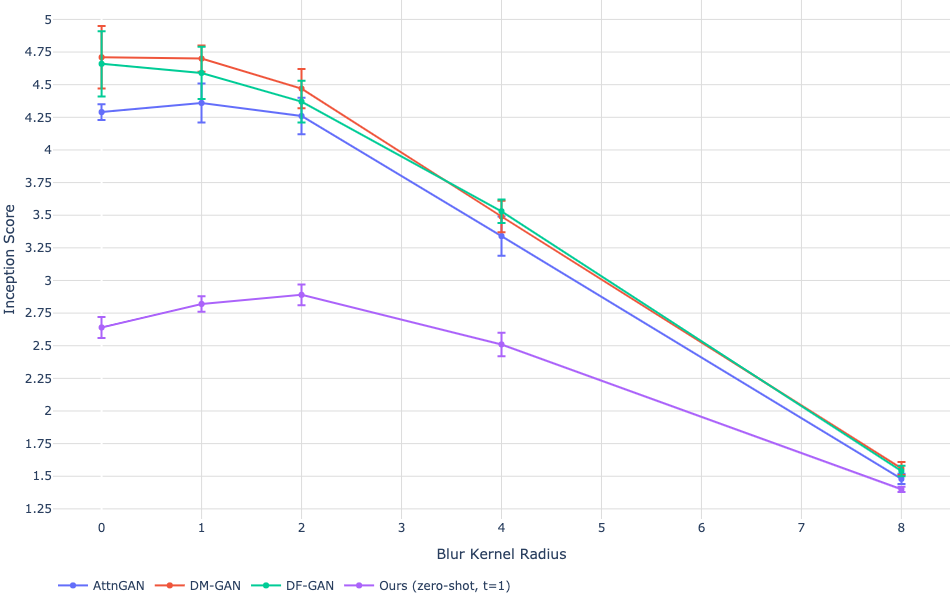}%
        \end{tabular}%
    }\hspace{2mm}
    \subfloat[FID and IS on MS-COCO as a function of the sample size used for reranking.]{%
        \begin{tabular}{@{}c}%
        \includegraphics[width=0.31\linewidth]{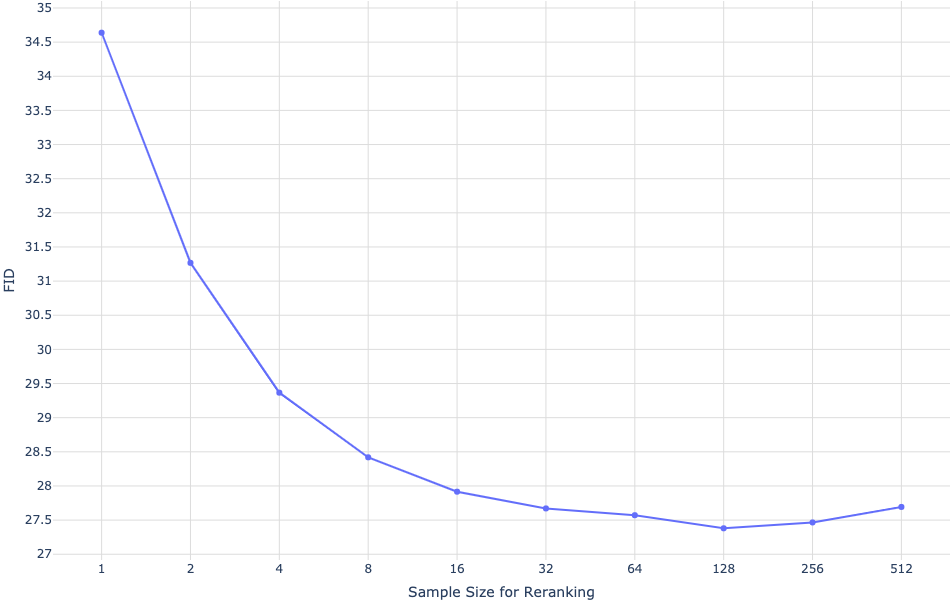} \\
        \includegraphics[width=0.31\linewidth]{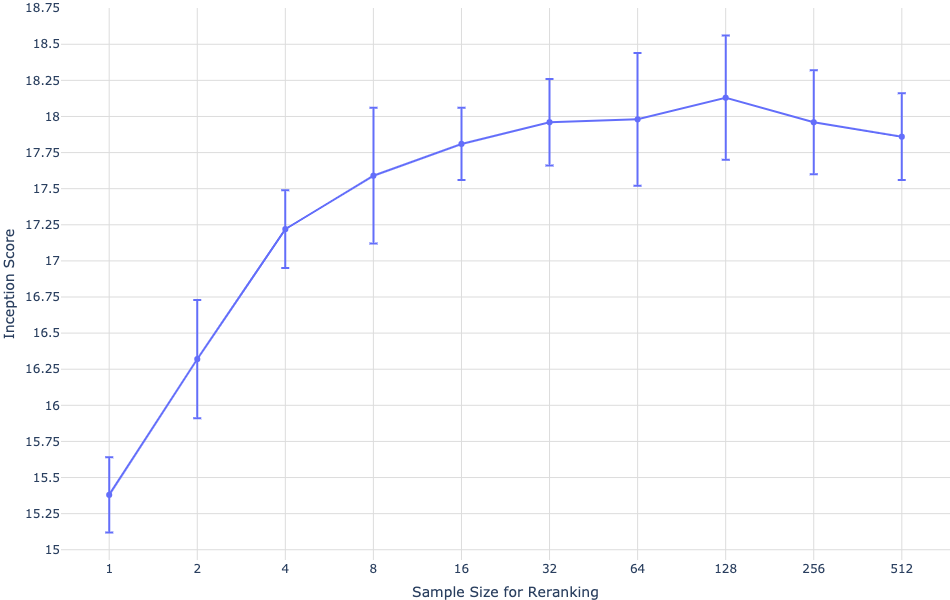}%
        \end{tabular}%
    }%
    \caption{Quantitative results on MS-COCO and CUB. Solid lines represent FID computed against the original validation sets, and dashed lines represent FID computed against validation sets with overlapping images removed (see Section~\ref{sec:data_overlap_analysis}). For MS-COCO, we evaluate all models on a subset of~\num{30000} captions sampled from the validation set. For CUB, we evaluate all models on all of the unique captions in the test set. We compute the FID and IS using the DM-GAN code, which is available at \url{https://github.com/MinfengZhu/DM-GAN}.}
    \label{fig:quant_results}
\end{figure*}

Figure~\ref{fig:quant_results}(a) shows that our model also obtains an FID score on MS-COCO within 2~points of the best prior approach, despite having never been trained on the captions. Our training data incorporates a filtered subset of YFCC100M, and we found that it includes about~$21\%$ of the images in the MS-COCO validation set from a de-duplication procedure described in the next section. To isolate this effect, we compute the FID statistics for the validation set both with these images (solid lines) and without them (dashed lines), finding no significant change in the results.

Training the transformer on the tokens from the dVAE encoder allows us to allocate its modeling capacity to the low-frequency information that makes images visually recognizable to us. However, it also disadvantages the model, since the heavy compression renders it unable to produce high-frequency details. To test the effect of this on the quantitative evaluations, we compute the~FID and~IS in Figure~\ref{fig:quant_results}(a) after applying a Gaussian filter with varying radius to both the validation images and samples from the models. Our approach achieves the best FID by a margin of about 6~points with a slight blur of radius~1. The gap between our approach and others tends to widen as the blur radius is increased. We also obtain the highest IS when the blur radius is greater than or equal to two.
\begin{figure}[t]
    \centering
    \includegraphics[width=\linewidth]{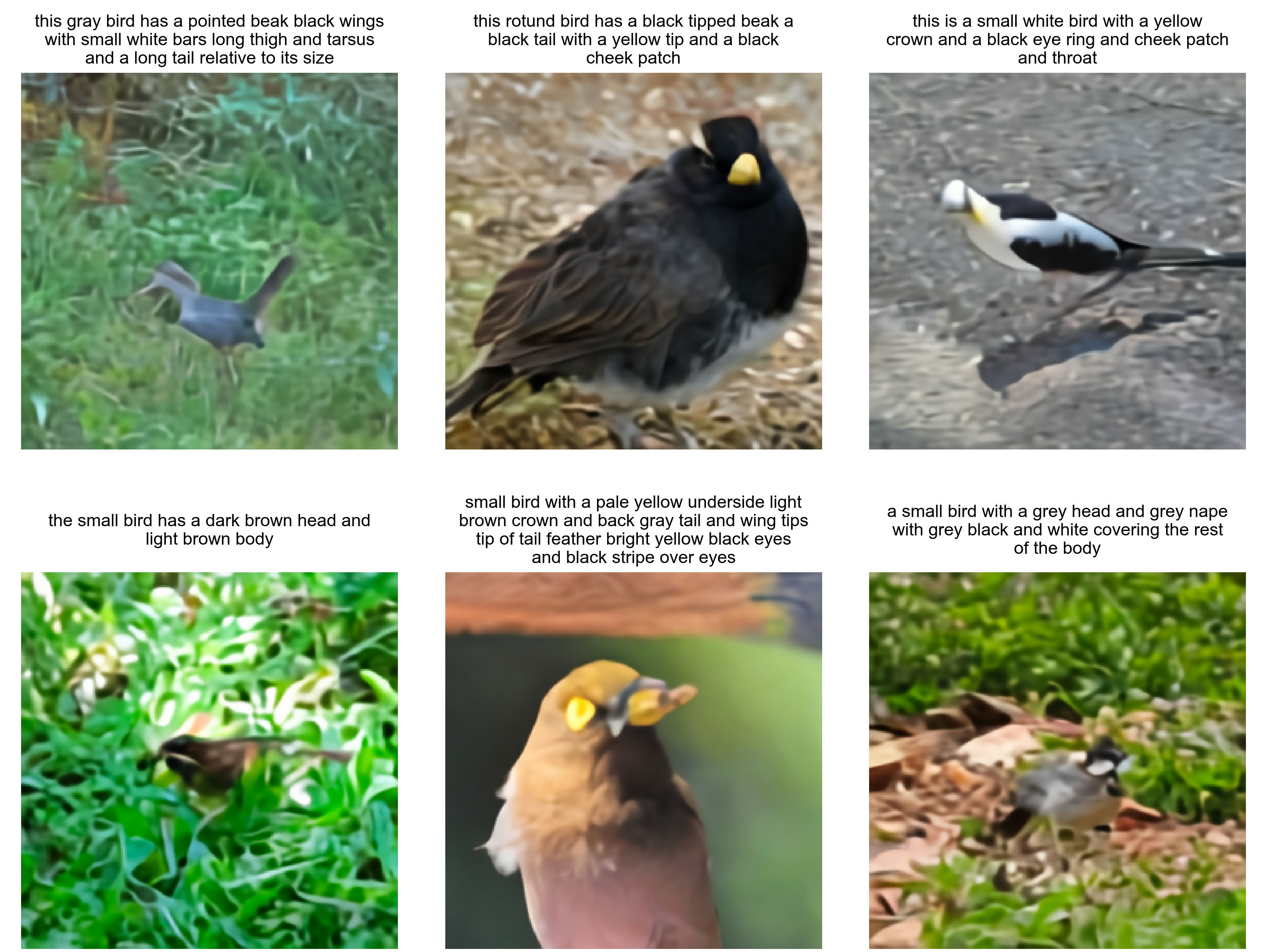}
    \caption{Zero-shot samples from our model on the CUB dataset.}
    \label{fig:cub_samples}
    \vspace{-1em}
\end{figure}

Our model fares significantly worse on the CUB dataset, for which there is a nearly 40-point gap in FID between our model and the leading prior approach~(Figure~\ref{fig:quant_results}(b)). We found an~$12\%$ overlap rate for this dataset, and again observed no significant difference in the results after removing these images. We speculate that our zero-shot approach is less likely to compare favorably on specialized distributions such as CUB. We believe that fine-tuning is a promising direction for improvement, and leave this investigation to future work. Samples from our model for captions in this dataset are shown in Figure~\ref{fig:cub_samples}.

Finally, Figure~\ref{fig:quant_results}(c) shows clear improvements in~FID and~IS for MS-COCO as the sample size used for reranking with the contrastive model is increased. This trend continues up to a sample size of~32, after which we observe diminishing returns.

\subsection{Data Overlap Analysis}
\label{sec:data_overlap_analysis}

We used the deduplication procedure described in~\citet{radford2021learning} to determine which images to remove. For each validation image, we find the closest image in the training data using a contrastive model specifically trained for this task. We then sort the images in descending order by closeness to their nearest matches in the training data. After inspecting the results by hand, we determine the images to remove by manually selecting a conservative threshold designed to minimize the false negative rate.

\subsection{Qualitative Findings}
\label{sec:qual_findings}

We found that our model has the ability to generalize in ways that we did not originally anticipate. When given the caption ``a tapir made of accordion...'' (Figure~\ref{fig:blog_samples1}), the model appears to draw a tapir with an accordion for a body, or an accordion whose keyboard or bass are in the shape of a tapir's trunk or legs. This suggests that it has developed a rudimentary ability to compose unusual concepts at high levels of abstraction.

Our model also appears to be capable of combinatorial generalization, such as when rendering text (Figure~\ref{fig:blog_samples2}) or when probed on sentences like ``an illustration of a baby hedgehog in a christmas sweater walking a dog'' (Figure~\ref{fig:blog_samples3}). Prompts like the latter require the model to perform variable binding~\cite{smolensky1990tensor,greff2020binding} -- it is the hedgehog that is in the christmas sweater, not the dog. We note, however, that the model performs inconsistently on the task, sometimes drawing both animals with christmas sweaters, or drawing a hedgehog walking a smaller hedgehog.

To a limited degree of reliability, we also find our model to be capable of zero-shot image-to-image translation controllable by natural language (Figure~\ref{fig:blog_samples4}). When the model is given the caption ``the exact same cat on the top as a sketch at the bottom'' and the top $15 \times 32$ part of the image token grid for a photo of a cat, it is able to draw a sketch of a similar looking cat on the bottom. 

This works with several other kinds of transformations, including image operations (e.g., changing the color of the image, converting it to grayscale, or flipping it upside-down) and style transfer (e.g., drawing the cat on a greeting card, a postage stamp, or a cell phone case). Some transformations, such as those that involve only changing the color of the animal, suggest that the model is capable of performing a rudimentary kind of object segmentation. We provide additional examples of zero-shot image-to-image translation in Section~\ref{sec:im2im}.

\section{Conclusion}

We investigate a simple approach for text-to-image generation based on an autoregressive transformer, when it is executed at scale. We find that scale can lead to improved generalization, both in terms of zero-shot performance relative to previous domain-specific approaches, and in terms of the range of capabilities that emerge from a single generative model. Our findings suggest that improving generalization as a function of scale may be a useful driver for progress on this task.


\section*{Acknowledgements}

We would like to thank Matthew Knight for reviewing the code release for this work, and Rewon Child, John Schulman, Heewoo Jun, and Prafulla Dhariwal for helpful early feedback on the paper. We would also like to thank Jong Wook Kim for writing the PyTorch package for the contrastive model described in \citet{radford2019language} that we used to rerank the samples from our model.

\bibliography{main}
\bibliographystyle{icml2020}

\newpage\appendix\onecolumn
\section{Details for Discrete VAE}

\subsection{Architecture}
\label{sec:dvae_arch}

The dVAE encoder and decoder are convolutional~\cite{lecun1998gradient} ResNets~\cite{he2016identity} with bottleneck-style resblocks. The models primarily use~$3 \times 3$ convolutions, with~$1 \times 1$ convolutions along skip connections in which the number of feature maps changes between the input and output of a resblock. The first convolution of the encoder is~$7 \times 7$, and the last convolution of the encoder (which produces the~$32 \times 32 \times 8192$ output used as the logits for the categorical distributions for the image tokens) is~$1 \times 1$. Both the first and last convolutions of the decoder are~$1 \times 1$. The encoder uses max-pooling (which we found to yield better ELB than average-pooling) to downsample the feature maps, and the decoder uses nearest-neighbor upsampling. The precise details for the architectures are given in the files \texttt{dvae/encoder.py} and \texttt{dvae/decoder.py} of the code release.

\subsection{Training}
\label{sec:dvae_train}
\begin{lstlisting}[language=Python,basicstyle=\footnotesize\ttfamily,caption={TensorFlow~\cite{abadi2016tensorflow} image preprocessing code for training dVAE. We use \texttt{target\_res = 256} and \texttt{channel\_count = 3}.},label={lst:dvae_preproc},captionpos=b,float=tp,floatplacement=tbp]
def preprocess_image(img, target_res):
    h, w  = tf.shape(img)[0], tf.shape(img)[1]
    s_min = tf.minimum(h, w)
    img   = tf.image.random_crop(img, 2 * [s_min] + [3])

    t_min = tf.minimum(s_min, round(9 / 8 * target_res))
    t_max = tf.minimum(s_min, round(12 / 8 * target_res))
    t     = tf.random.uniform([], t_min, t_max + 1, dtype=tf.int32)
    img   = tf.image.resize_images(img, [t, t], method=tf.image.ResizeMethod.AREA,
                align_corners=True)
    img   = tf.cast(tf.rint(tf.clip_by_value(img, 0, 255)), tf.uint8)
    img   = tf.image.random_crop(img, 2 * [target_res] + [channel_count])
    return tf.image.random_flip_left_right(img)
\end{lstlisting}
The dVAE is trained on the same dataset as the transformer, using the data augmentation code given in Listing~\ref{lst:dvae_preproc}. Several quantities are decayed during training, all of which use a cosine schedule:
\begin{enumerate}
    \item The KL weight~$\beta$ is increased from~$0$ to~$6.6$ over the first~\num{5000} updates. \citet{bowman2015generating} use a similar schedule based on the sigmoid function.
    \item The relaxation temperature~$\tau$ is annealed from~$1$ to~$1 / 16$ over the first~\num{150000} updates. Using a linear annealing schedule for this typically led to divergence.
    \item The step size is annealed from~$1 \cdot 10^{-4}$ to~$1.25 \cdot 10^{-6}$ over~\num{1200000} updates.
\end{enumerate}
The decay schedules for the relaxation temperature and the step size are especially important for stability and successful optimization.

We update the parameters using AdamW~\cite{loshchilov2017decoupled} with $\beta_1=0.9$, $\beta_2=0.999$, $\epsilon=10^{-8}$, and weight decay multiplier~$10^{-4}$. We use exponentially weighted iterate averaging for the parameters with decay coefficient~$0.999$. The reconstruction term in the~ELB is a joint distribution over the~$256 \times 256 \times 3$ values for the image pixels, and the KL term is a joint distribution over the~$32 \times 32$ positions in the spatial grid output by the encoder. We divide the overall loss by~$256 \times 256 \times 3$, so that the weight of the KL term becomes~$\beta / 192$, where~$\beta$ is the KL weight. The model is trained in mixed-precision using standard (i.e., global) loss scaling on~$64$ 16~GB NVIDIA V100 GPUs, with a per-GPU batch size of~$8$, resulting in a total batch size of~512. It is trained for a total of~\num{3000000} updates.

\subsection{The Logit-Laplace Distribution}
\label{sec:dvae_logit_laplace}

The~$\ell_1$ and~$\ell_2$ reconstruction objectives are commonly used when training VAEs. These objectives correspond to using Laplace and Gaussian distributions for~$\ln p_\theta(x \given y, z)$ in Equation~\ref{eq:elb}, respectively. There is a strange mismatch in this modeling choice: pixel values lie within a bounded interval, but both of these distributions are supported by the entire real line. Hence, some amount of likelihood will be placed outside the admissible range of pixel values.

We present a variant of the Laplace distribution that is also supported by a bounded interval. This resolves the discrepancy between the range of the pixel values being modeled and the support of the distribution used to model them. We consider the pdf of the random variable obtained by applying the sigmoid function to a Laplace-distributed random variable. This pdf is defined on~$(0, 1)$ and is given by
\begin{equation}
    f(x \given \mu, b) = \frac{1}{2b x (1 - x)} \exp\left(-\frac{|\operatorname{logit}(x) - \mu|}{b}\right);
    \label{eq:logit_laplace_pdf}
\end{equation}
we call it the \emph{logit-Laplace distribution.} We use the logarithm of the RHS of Equation~\ref{eq:logit_laplace_pdf} as the reconstruction term for the training objective of the dVAE.

The decoder of the dVAE produces six feature maps representing the sufficient statistics of the logit-Laplace distribution for the RGB channels of the image being reconstructed. The first three feature maps represent the~$\mu$ parameter for the RGB channels, and the last three represent~$\ln b$. Before feeding an image into the dVAE encoder, we transform its values using~$\varphi: [0, 255] \to (\epsilon, 1 - \epsilon)$, which is given by
\begin{equation}
    \varphi : x \mapsto \frac{1 - 2\epsilon}{255} x + \epsilon.
\end{equation}
This restricts the range of the pixel values to be modeled by the dVAE decoder to~$(\epsilon, 1 - \epsilon)$, which avoids numerical problems arising from the~$x (1 - x)$ in Equation~\ref{eq:logit_laplace_pdf}. We use~$\epsilon = 0.1$. To reconstruct an image for manual inspection or computing metrics, we ignore~$\ln b$ and compute~$\hat{x} = \varphi^{-1}(\operatorname{sigmoid}(\mu))$, where~$\mu$ is given by the first three feature maps output by the dVAE decoder.\footnote{See \texttt{notebooks/usage.ipynb} of the code release for an example.}

\section{Details for Transformer}

\subsection{Architecture}
\label{sec:xf_arch}
\begin{figure}[t]
    \centering
    \includegraphics[width=\textwidth]{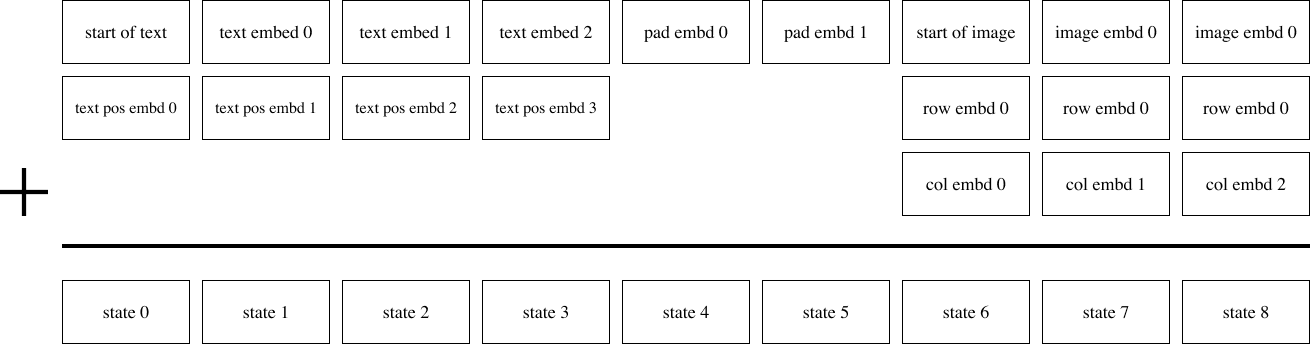}
    \caption{Illustration of the embedding scheme for a hypothetical version of our transformer with a maximum text length of 6~tokens. Each box denotes a vector of size~$d_\mathrm{model} = 3968$. In this illustration, the caption has a length of 4~tokens, so 2~padding tokens are used (as described in Section~\ref{sec:learning_prior}). Each image vocabulary embedding is summed with a row and column embedding.}
    \label{fig:xf_embds}
\end{figure}
\begin{figure}[t]
    \centering
    \subfloat[Row attention mask.]{%
        \includegraphics[width=0.24\linewidth]{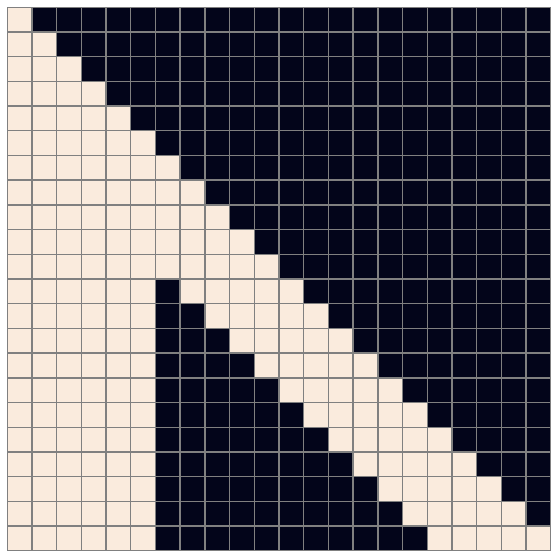}%
    }%
    \hspace{2mm}%
    \subfloat[Column attention mask.]{%
        \includegraphics[width=0.24\linewidth]{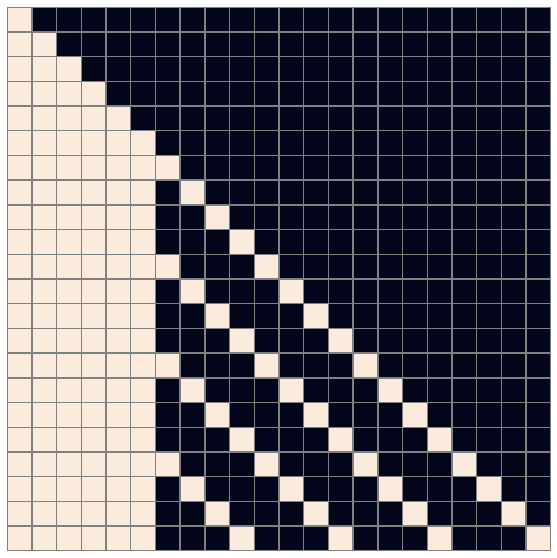}%
    }%
    \hspace{2mm}%
    \subfloat[Column attention mask with transposed image states.]{%
        \includegraphics[width=0.24\linewidth]{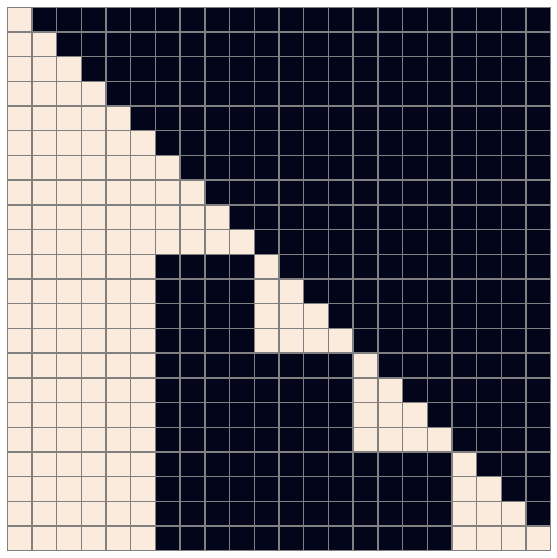}%
    }%
    \hspace{2mm}%
    \subfloat[Convolutional attention mask.]{%
        \includegraphics[width=0.24\linewidth]{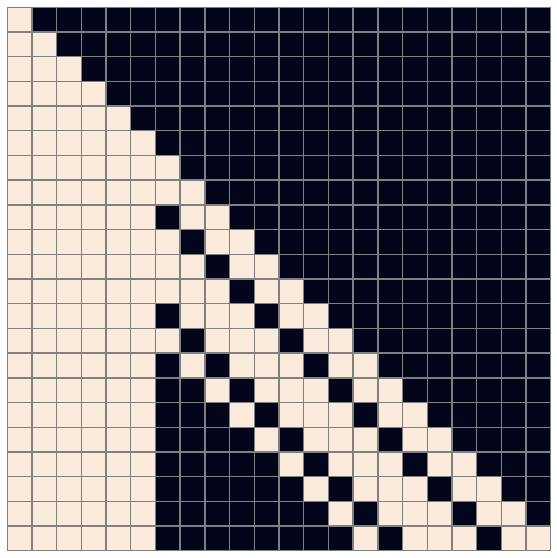}%
    }%
    \caption{Illustration of the three types of attention masks for a hypothetical version of our transformer with a maximum text length of 6~tokens and image length of 16~tokens (i.e., corresponding to a~$4 \times 4$ grid). Mask~(a) corresponds to row attention in which each image token attends to the previous~5 image tokens in raster order. The extent is chosen to be~5, so that the last token being attended to is the one in the same column of the previous row. To obtain better GPU utilization, we transpose the row and column dimensions of the image states when applying column attention, so that we can use mask~(c) instead of mask~(b). Mask~(d) corresponds to a causal convolutional attention pattern with wraparound behavior (similar to the row attention) and a~$3 \times 3$ kernel. Our model uses a mask corresponding to an~$11 \times 11$ kernel.}
    \label{fig:xf_attn}
\end{figure}
Our model is a decoder-only sparse transformer of the same kind described in \citet{child2019generating}, with broadcasted row and column embeddings for the part of the context for the image tokens. A complete description of the embedding scheme used in our model is shown in Figure~\ref{fig:xf_embds}. We use 64~attention layers, each of which uses 62~attention heads with a per-head state size of~64.

The model uses three kinds of sparse attention masks, which we show in Figure~\ref{fig:xf_attn}. The convolutional attention mask (Figure~\ref{fig:xf_attn}(d)) is only used in the last self-attention layer. Otherwise, given the index~$i$ of a self-attention layer (with~$i \in [1, 63]$), we use the column attention mask (Figure~\ref{fig:xf_attn}(c)) if $i - 2 \!\!\mod 4 = 0$, and row attention otherwise. E.g., the first four self-attention layers use ``row, column, row, row'', respectively. With the exception of the convolutional attention mask, which we found to provide a small boost in performance over the row and dense causal attention masks when used in the final self-attention layer, this is the same configuration used in~\citet{child2019generating}.

\subsection{Training}
\label{sec:xf_train}
\begin{lstlisting}[language=Python,basicstyle=\footnotesize\ttfamily,caption={TensorFlow~\cite{abadi2016tensorflow} image preprocessing code for training the transformer. We use \texttt{target\_res = 256} and \texttt{channel\_count = 3}.},label={lst:xf_preproc},captionpos=b,float=tp,floatplacement=tbp]
def preprocess_image(img, target_res):
    h, w  = tf.shape(img)[0], tf.shape(img)[1]
    s_min = tf.minimum(h, w)

    off_h = tf.random.uniform([], 3 * (h - s_min) // 8,
        tf.maximum(3 * (h - s_min) // 8 + 1, 5 * (h - s_min) // 8),
        dtype=tf.int32)
    off_w = tf.random.uniform([], 3 * (w - s_min) // 8,
        tf.maximum(3 * (w - s_min) // 8 + 1, 5 * (w - s_min) // 8),
        dtype=tf.int32)

    # Random full square crop.
    img   = tf.image.crop_to_bounding_box(img, off_h, off_w, s_min, s_min)
    t_max = tf.minimum(s_min, round(9 / 8 * target_res))
    t     = tf.random.uniform([], target_res, t_max + 1, dtype=tf.int32)
    img   = tf.image.resize_images(img, [t, t], method=tf.image.ResizeMethod.AREA,
                align_corners=True)
    img   = tf.cast(tf.rint(tf.clip_by_value(img, 0, 255)), tf.uint8)

    # We don't use hflip aug since the image may contain text.
    return tf.image.random_crop(img, 2 * [target_res] + [channel_count])
\end{lstlisting}

When training the transformer, we apply data augmentation to the images before encoding them using the dVAE encoder. We use slightly different augmentations from the ones used to train the dVAE; the code used for this is given in Listing~\ref{lst:xf_preproc}. We also apply 10\% BPE dropout when BPE-encoding the captions for training. The model is trained using per-resblock scaling (see Section~\ref{sec:mp_train}) and gradient compression (see Section~\ref{sec:dist_opt}) with total compression rank~896 (so that each GPU uses a compression rank of~112 for its parameter shards). As shown in Table~\ref{tab:cmp_rank}, this results in a compression rate of about~86\%, which we analyze in Section~\ref{sec:dist_train_bw}.

We update the parameters using AdamW with~$\beta_1 = 0.9$, $\beta_2 = 0.96$, $\epsilon = 10^{-8}$, and weight decay multiplier~$4.5 \cdot 10^{-2}$. We clip the decompressed gradients by norm using a threshold of~4, prior to applying the Adam update. Gradient clipping is only triggered during the warm-up phase at the start of training. To conserve memory, most Adam moments~(see Section~\ref{sec:mp_train_guidelines} for details) are stored in 16-bit formats, with a 1-6-9~format for the running mean (i.e., 1~bit for the sign, 6~bits for the exponent, and 9~bits for the significand), and a 0-6-10~format for the running variance. We clip the estimate for running variance by value to~5 before it is used to update the parameters or moments. Finally, we apply exponentially weighted iterate averaging by asynchronously copying the model parameters from the~GPU to the~CPU once every 25~updates, using a decay coefficient of~0.99.

We trained the model using 1024, 16~GB NVIDIA V100 GPUs and a total batch size of~$1024$, for a total of~\num{430000} updates. At the start of training, we use a linear schedule to ramp up the step size to~$4.5 \cdot 10^{-4}$ over~\num{5000} updates, and halved the step size each time the training loss appeared to plateau. We did this a total of five times, ending training with a final step size that was 32~times smaller than the initial one. We reserved about~\num{606000} images for validation, and did not observe overfitting at any point during training.

\section{Details for Data Collection}
\label{sec:data_collection_details}

In order to train the 12-billion parameter transformer, we created a dataset of a similar scale to JFT-300M by collecting 250~million text-image pairs from the internet. As described in Section~\ref{sec:data_collection}, this dataset incorporates Conceptual Captions, the text-image pairs from Wikipedia, and a filtered subset of YFCC100M. We use a subset of the text, image, and joint text and image filters described in \citet{sharma2018conceptual} to construct this dataset. These filters include discarding instances whose captions are too short, are classified as non-English by the Python package \texttt{cld3}, or that consist primarily of boilerplate phrases such as ``photographed on \texttt{<date>}'', where \texttt{<date>} matches various formats for dates that we found in the data. We also discard instances whose images have aspect ratios not in~$[1 / 2, 2]$. If we were to use to very tall or wide images, then the square crops used during training would likely exclude objects mentioned in the caption.

\section{Guidelines for Mixed-Precision Training}
\label{sec:mp_train_guidelines}
\begin{figure}[t]
    \centering
    \includegraphics[width=0.4\textwidth]{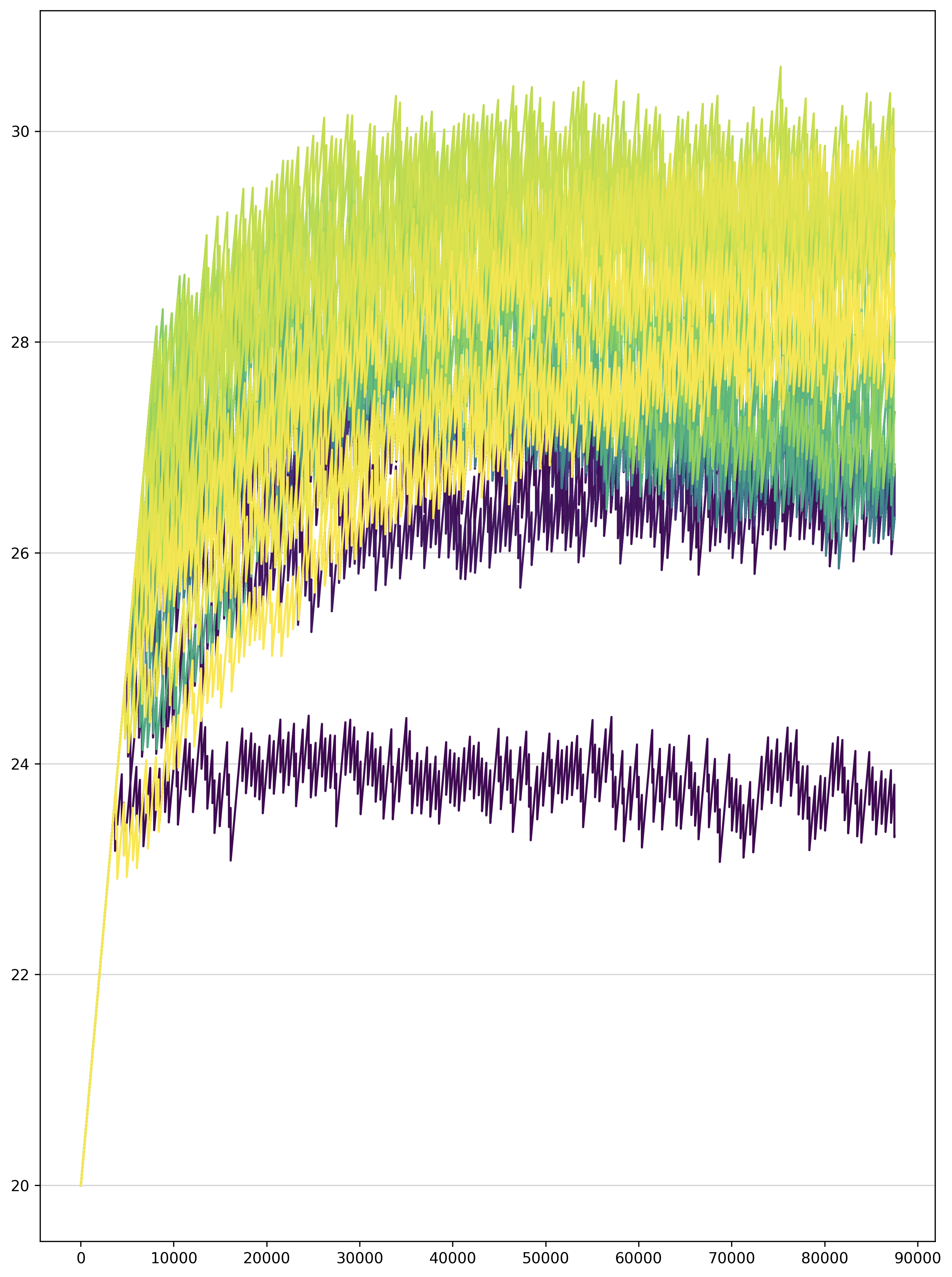}
    \caption{Plot of per-resblock gradient scales for a 2.8-billion parameter text-to-image transformer trained without gradient compression. The $x$-axis is parameter updates, and the $y$-axis is the base-2 logarithm of the gradient scale. Darkest violet corresponds to the first resblock, and brightest yellow corresponds to the last (of which there are 128 total). The gradient scale for the second MLP resblock hovers at around~$2^{24}$, while the others stay within a 4-bit range. The extent of this range increases as the model is made larger.}
    \label{fig:grad_scale_plot}
\end{figure}
The most challenging part of this project was getting the model to train in 16-bit precision past one billion parameters. We were able to do this after detecting for underflow in various parts of training, and revising the code to eliminate it. We developed a set of guidelines as a result of this process that we present here.\footnote{Fewer of these guidelines may be necessary on hardware like the TPU that has native support for the bfloat16 format, since the larger 8-bit exponent range makes underflow less likely to occur.}
\begin{enumerate}
    \item \textbf{Use per-resblock gradient scaling (Figure~\ref{fig:grad_scaling}) instead of standard loss scaling.} Our model uses 128~gradient scales, one for each of its resblocks. All of the gradient scales are initialized to~$M \cdot 2^{13}$, where~$M$ is the number of data-parallel replicas (i.e., the number of GPUs). In our setup, each grad scale is multiplied by~$2^{1 / 1000}$ at every parameter update when there are no nonfinite values for any parameter gradient in that resblock. Otherwise, we divide the grad scale by~$\sqrt{2}$ and skip the update. We also disallow consecutive divisions of the same grad scale within a window of $125$~updates. All grad scales are clamped to the range~$[M \cdot 2^7, M \cdot 2^{24}]$ after being updated. Figure~\ref{fig:grad_scale_plot} shows the gradient scales in the early phase of training for a 2.8-billion parameter model.
    \item \textbf{Only use 16-bit precision where it is really necessary for performance.} In particular, store all gains, biases, embeddings, and unembeddings in 32-bit precision, with 32-bit gradients (including for remote communication) and 32-bit Adam moments. We disable gradient compression for these parameters (though PowerSGD would not make sense for 1D parameters like gains and biases). The logits for the text and image tokens are computed and stored in 32-bit precision. We found that storing the embeddings in 16-bit precision sometimes caused divergence early in optimization, and using 16-bit logits resulted in a small shift in the training curve, so we switched to use 32-bit precision out of an abundance of caution.
    \item \textbf{Avoid underflow when dividing the gradient.} For data-parallel training, we need to divide the gradients by the total number of data-parallel workers~$M$. One way to do this is to divide the loss by the per-machine batch size, and then divide the parameter gradients by~$M$ before summing them over the machines (using all-reduce). To save time and space, the gradients are usually computed and stored in 16-bit precision. When~$M$ is large, this division could result in underflow before the gradients are summed. On the other hand, if we attempt to sum the gradients first and then divide them later, we could encounter overflow in the all-reduce.
    
    Our solution for this problem attempts to minimize the loss of information in the division prior to the all-reduce, without danger of overflow. To do this, we divide the loss by the overall batch size (which includes~$M$ as a factor) rather than the per-machine batch size, and multiply the gradient scales by~$M$ to compensate, as described in~(1). Then, prior to the all-reduce operation, we divide the gradients by a constant that was tuned by hand to avoid both underflow and overflow. This was done by inspecting histograms of the exponents (i.e., base-2 logarithms) of the absolute values of the scalar components of the per-parameter gradients. Since the gradient scaling keeps the gradients close to right end of the exponent range of the 16-bit format, we found that the same constant worked well for all parameters in the model with 16-bit gradients. When using PowerSGD, we chose different constants for the~$P$ and~$Q$ matrices.
\end{enumerate}

\section{Details for Distributed Optimization}

We use PowerSGD~\cite{vogels2019powersgd} to compress the gradients with respect to all parameters except the embeddings, unembeddings, gains, and biases. In Section~\ref{sec:dist_train_bw}, we derive an expression for the reduction in the amount of data communicated as a function of the compression rank and model size. In Section~\ref{sec:dist_train_impl}, we present a detailed overview of our adaptation of PowerSGD, and the modifications we had to make in order to fix performance regressions, some of which only manifest at billion-parameter scale.

\subsection{Bandwidth Analysis}
\label{sec:dist_train_bw}
\begin{table}[]
    \centering\small
    \begin{tabular}{cccc}
        \toprule
         Parameter Names & Parameter Shard Gradient Shape (No Compression) & $P$ shape & $Q$ shape \\
         \midrule
         qkv and post-attention matrices & $d \times (d / m)$ & $d \times (r / m)$ & $(r / m) \times (d / m)$ \\
         First MLP matrix & $d \times (4d / m)$ & $d \times (r / m)$ & $(r / m) \times (4d / m)$ \\
         Second MLP matrix & $(4d / m) \times d$ & $(4d / m) \times (r / m)$ & $(r / m) \times d$ \\
         Total size & $12d^2 / m$ & $(5drm + 4dr) / m^2$ & $(drm + 8dr) / m^2$ \\
         \bottomrule
    \end{tabular}
    \caption{We analyze the amount of data sent from each GPU on a given machine to GPUs on other machines, in the case where we shard the parameters among the~$m$ GPUs on each machine. Here, $r$ denotes the rank used for compression, and $d$ the transformer hidden size. The compression ratio is given by the sum of the last two columns of the last row, divided by the first column of the last row. This comes out to $r(m + 2) / (2dm)$, which for $m = 8$ is $5r / 8d$.}
    \label{tab:cmp_ratio}
\end{table}
Gradient compression uses the factorization $G \approx P Q^t$, where~$P$ and~$Q$ both have rank $r$. Instead of using a single all-reduce to transmit $G$, we use two, smaller all-reduces to transmit both~$P$ and~$Q^t$ in succession. Hence, the compression ratio is the sum of the sizes of the~$P$ and~$Q$ matrices divided by the sum of the sizes of the~$G$ matrices. We shard along axis~1 for all parameters except for the second MLP matrix. The derivation of the compression ratio in our setup is given in Table~\ref{tab:cmp_ratio}. We note that the choice of shard axis changes the compression ratio for the MLP matrices. Finally, this analysis excludes the embeddings, unembeddings, gains, and biases, for which we do not use compression. The total fraction of the bandwidth used by these parameters becomes smaller as the model size is increased.

\subsection{Implementation Details}
\label{sec:dist_train_impl}

We describe the steps in our implementation of PowerSGD in detail, since these details were crucial in getting it to work efficiently and reliably at billion-parameter scale.
\begin{enumerate}
    \item Our training setup uses a combination of parameter sharding and gradient compression, as described in Section~\ref{sec:dist_opt}. During backpropagation, while recomputing the activations and computing the gradients for the current resblock, we prefetch the parameters for the preceding resblock using all-gather. Once each GPU has computed the gradient with respect to a full parameter matrix, we compute the average of the slice of the gradient corresponding to the GPU's parameter shard, and discard the full gradient immediately to conserve memory. This average is taken over all of the GPUs on a machine using reduce-scatter.
    \item If there are no nonfinite values in the result of the reduce-scatter~(which could be caused by overflow in backpropagation or the reduce-scatter), we divide the result by the resblock's gradient scale, and add it to the error buffer (i.e., the buffer used for error correction). Otherwise, we do nothing and proceed with backpropagation; a single nonfinite value in the gradient means that the entire update will be skipped, which happens about~5\% of the time. The error buffer uses the same 1-6-9 format used for the Adam mean, which we describe in Section~\ref{sec:xf_train}; the larger exponent range ensures that this division does not result in underflow. Adding the gradients directly to the error buffers avoids redundantly allocating another set of buffers of size equal to the parameter shard gradients.
    \item Once the reduce-scatter operations for the resblock have finished, we schedule the operations to compute the~$P$ matrices from the errors buffers and the~$Q$ matrices, whose values are fixed at the start of training (see Section~\ref{sec:dist_opt}). Both the~$P$ and~$Q$ matrices are stored in 1-6-9 format and have their values scaled by predetermined constants, as discussed in Section~\ref{sec:mp_train_guidelines}.
    \item Once each GPU has computed the~$P$ matrices for the parameter shards in a resblock, they are averaged with the~$P$ matrices from the GPUs with the same ordinal on all other machines, using a single, grouped all-reduce operation. This all-reduce is carried out in the 1-6-9 format, using a custom kernel. The grouping results in better bandwidth utilization, since it avoids scheduling many all-reduce calls for smaller, individual parameters, each of which carries some overhead. We clamp any infinities in the results of the all-reduce to the maximum value of the 1-6-9 format (which is slightly less than 16), retaining the sign. With our choice of scaling factors for the~$P$ and~$Q$ matrices, this clamping happens very rarely.
    \item Once the all-reduce operation for the~$P$ matrices for a resblock have finished, we orthogonalize the columns of the resulting matrices. We use a custom Householder orthogonalization kernel rather than Gram-Schmidt, as we found the latter to be numerically unstable. We also add~$\epsilon I_{m \times r}$ to~$P$ in order to ensure that the result is not near rank-deficient, where~$\epsilon = 10^{-6}$. Here, $I_{m \times r}$ is a rectangular matrix of the same size as the~$P$ matrix to which it is added; it contains the~$r \times r$ identity matrix and has zeros elsewhere. The orthogonalizalied~$P$ matrices are stored in 1-6-9 format, but without scaling. 
    \item Once the~$P$ matrices for a resblock have been orthogonalized, we schedule the operations to compute the new~$Q$ matrices from the error buffers and the~$P$ matrices.
    \item Once the new~$Q$ matrices for a resblock have been computed, we schedule another grouped all-reduce, similar to what we did for the~$P$ matrices. As in step~(4), we clamp all infinities in the results of the all-reduce to the maximum value of the 1-6-9 format, retaining the sign. The error buffers for the resblock have now been decomposed into low-rank factors~$P$ and~$Q^t$.
    \item The gradients for all parameters that are not compressed are grouped together into a single, 32-bit precision all-reduce. Section~\ref{sec:mp_train_guidelines} explains why we use 32-bit precision for these parameters and their gradients.
    \item Once all GPUs on a machine have finished steps~(7) and~(8) for every resblock in the model, the values of the~$P$ and~$Q$ matrices for the same parameter shard on all machines will be identical. We then compute the global gradient norm, which is the sum of two quantities: (a)~the sum of the squared Frobenius norms of the~$Q$ matrices over all of the parameter shards on a machine, and (b)~the sum of the squared norms of the gradients for the parameter shards that do not use compression, taken over all such parameter shards on a machine. We need to compute this value for gradient clipping (see Section~\ref{sec:xf_train}).
    \item While computing the global norm, we also synchronize the information from step~(2) about which parameter shard gradients contained nonfinite values after the reduce-scatter. After doing this, we have two pieces of information for each parameter shard: (a) whether its error buffer from step~(2) contains nonfinite values on the current GPU, and (b) whether~$P$ or~$Q$ contains nonfinite values. We cannot rely on the values of the~$P$ and~$Q$ matrices to determine~(b), since we clamp infinities as described in step~(4). If we find that the gradient with respect to any parameter shard on the machine contains nonfinite values, then we set the global norm to infinity.
    \item Once all of the all-reduces have finished and the global norm has been computed, we can apply the parameter updates. Like backpropagation, the parameter updates proceed resblock-by-resblock. The first step is to compute the decompressed gradients by forming the product~$PQ^t$ for all parameters in a given resblock. To avoid overflow, these products are computed in 32-bit precision. We can then apply the Adam update to the parameters using the decompressed gradients and the global norm computed in step~(9). If the global norm is not finite, then the update to the parameters and Adam moments is skipped. We note that the decompressed gradient must be divided by the scale of the $Q$~matrix (the $P$~matrix is stored without scaling after orthogonalization).
    \item The second step is the update to the error buffers. First, we use the results from step~(10) to check if the~$P$ and~$Q$ matrices for a given parameter shard contain only finite values. If this is the case, then we divide the decompressed gradient by the total number of machines, and subtract it from the current value for the error buffer. This sets the error buffer to the difference between the ``local'' gradient averaged over the GPUs on the machine using reduce-scatter, and the ``remote'' decompressed gradient (i.e., the ``error''). If either~$P$ or~$Q$ contains nonfinite values, then we check if the error buffer computed in step~(2) contains only finite values. If it does, then we preserve its value and do nothing. If it does not, then we set it to zero. The purpose of this tedious logic is to set an error buffer to zero only when we must do so, because it has been contaminated with nonfinite values. We found that error buffers getting set to zero too frequently by gradient scaling events leads to performance regressions.
    \item The parameter shards whose gradients are not compressed are updated separately.
\end{enumerate}

We also note the following important optimizations:
\begin{enumerate}
    \item There are several opportunities for overlap between compute and communication in the above steps. For example, while we are running step~(2) for resblock~$i$, we can proceed to steps~(3)--(8) for all resblocks~$j > i$. Exploiting opportunities for overlap is necessary to achieve good performance.
    \item We throttle specific operations that are liable to exhaust all available memory. For example, we only prefetch the parameters from the preceding resblock when the reduce-scatter operations have finished for the current one. Otherwise, we risk running out of memory by holding on to the full parameters. We also throttle the Adam updates, so that we do not decompress all of the gradients at once.
    \item There are two places in the implementation where the transposition matters: (a)~the choice of shard axis for the MLP matrices and (b)~whether we compute the low-rank factorization for a gradient or its transpose. The former influences the bandwidth analysis, which we present in Section~\ref{sec:dist_train_bw}. The latter influences the cost of the orthogonalization. Suppose that the gradient~$G$ is~$m \times n$ and its low-rank factors~$P$ and~$Q^t$ are~$m \times r$ and~$r \times n$, respectively, with~$r \ll m, n$. To make orthogonalization cheaper, we transpose~$G$ appropriately so that~$m \leqslant n$.
    
    At first glance, it may seem like a limitation that the NCCL all-gather and reduce-scatter primitives shard along axis~0 only. We may need to transpose some matrices before and after communication operations because of~(a) and~(b), which would require additional time and potentially special care to avoid out-of-memory errors. In fact, we never actually needed to do this. This is because we stored some of the parameters in their transposed formats and exploited the \texttt{transpose\_a} and \texttt{transpose\_b} parameters of the matrix multiplication kernels used in forward propagation, backpropagation, and steps (1)--(13) above. This allowed us to avoid explicit transposition while retaining the freedom to choose how to handle~(a) and~(b).
    \item In step~(12) above, we note that setting the error buffers to zero too often can cause performance regressions. We wanted to avoid doing this when resuming training from a checkpoint, which happens more frequently for larger jobs as it is likely that a machine will periodically fail. Naively, this would require uploading the error buffers from all of the machines along with the model checkpoints. Since we use a total of 128~machines for training, this would lead to 128~times greater storage usage, which is extremely wasteful.
    
    Fortunately, this is unnecessary, as error correction depends only on the sum of the error buffers. This property follows from linearity and the sequence of operations used by PowerSGD. Hence, it suffices to store the sums of the errors buffers taken across all GPUs with the same ordinal. When resuming from a checkpoint, we can divide the error buffers by the total number of machines and broadcast them.
\end{enumerate}

\section{Details for Human Evaluation Experiments}
\label{sec:human_eval}
\begin{figure*}[t]
    \centering
    \includegraphics[width=\linewidth]{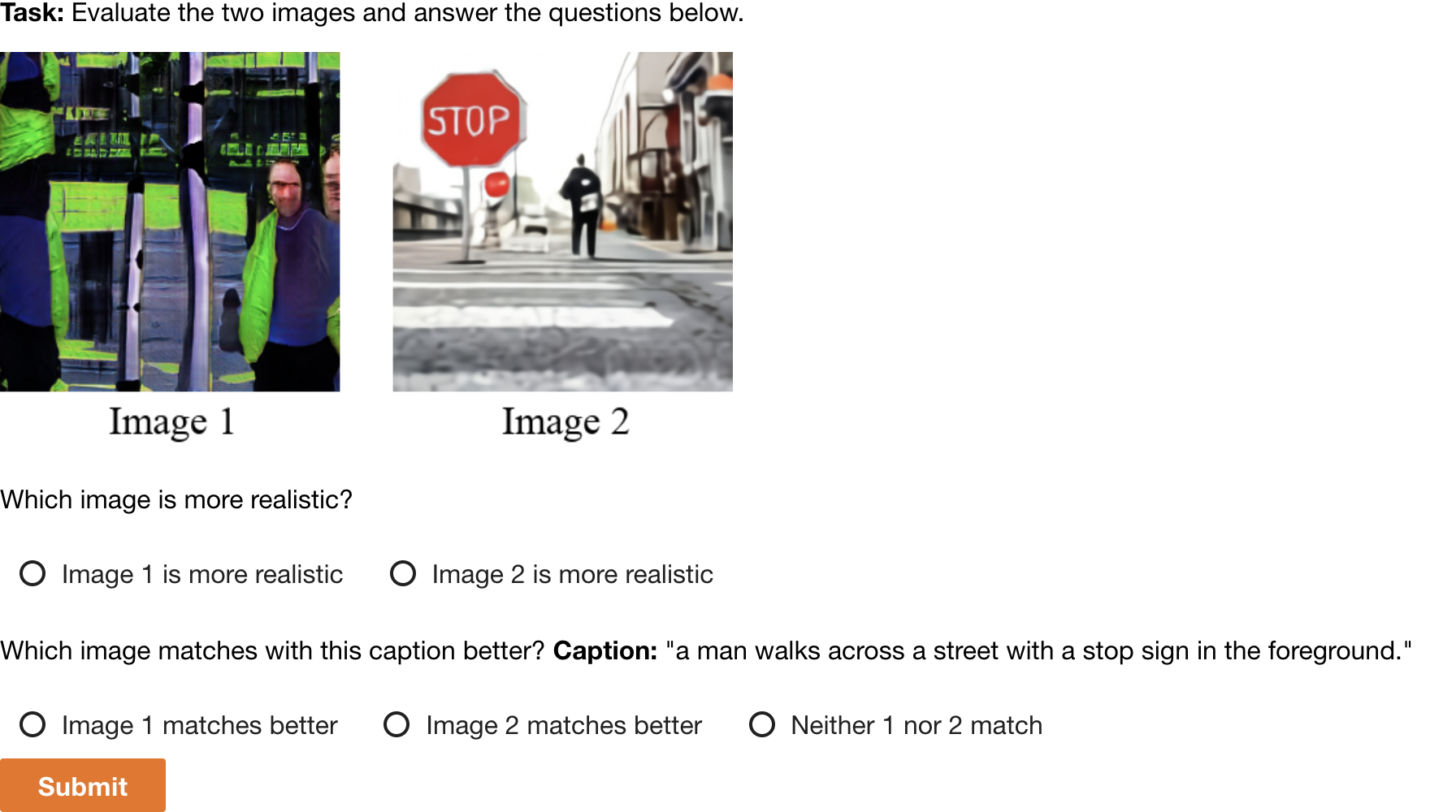}
    \caption{Example task interface shown to workers.}
    \label{fig:example_human_evals_task}
\end{figure*}
We start with a list of~\num{1000} captions and generate one sample image per model per caption. Captions and sample images are then used to create \num{1000} image comparison tasks per experiment, which we submitted to Amazon's Mechanical Turk. Each task was answered by five distinct workers. Workers were asked to compare two images and answer two questions about them:~(1) which image is most realistic, and (2)~which image best matches the shared caption. The experimental setup provided to workers is shown in Figure~\ref{fig:example_human_evals_task}. One worker's answers were disqualified due to a high rate of disagreement with other workers combined with a fast answer velocity (with many submission times under 4 seconds); all other worker answers were kept.

\section{Zero-Shot Image-to-Image Translation}
\label{sec:im2im}
\begin{figure*}
\centering
\captionsetup[subfigure]{width=1.9in}
\subfloat[``the exact same cat on the top as a sketch on the bottom'']{%
\setlength\extrarowheight{-3pt}
\begin{tabular}{@{}c@{\hskip 0.05in}c@{}}
  \fbox{\includegraphics[scale=0.27]{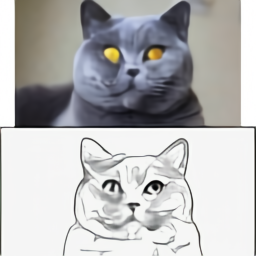}}
  & \fbox{\includegraphics[scale=0.27]{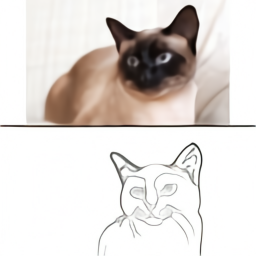}} 
 \\
  \fbox{\includegraphics[scale=0.27]{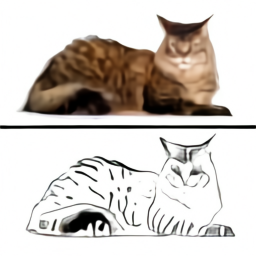}} 
 & \fbox{\includegraphics[scale=0.27]{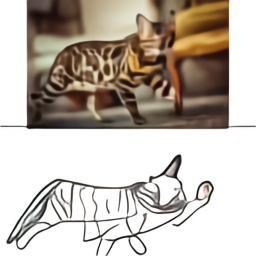}} 
 \\
\end{tabular}}%
\hspace{0.75mm}%
\subfloat[``the exact same photo on the top reflected upside-down on the bottom'']{%
\setlength\extrarowheight{-3pt}
\begin{tabular}{@{}c@{\hskip 0.05in}c@{}}
  \fbox{\includegraphics[scale=0.27]{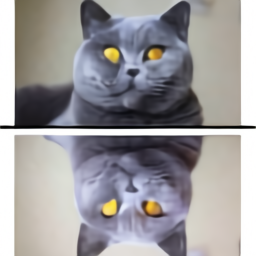}}
  & \fbox{\includegraphics[scale=0.27]{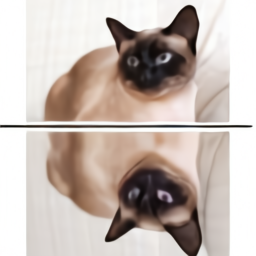}} 
 \\
  \fbox{\includegraphics[scale=0.27]{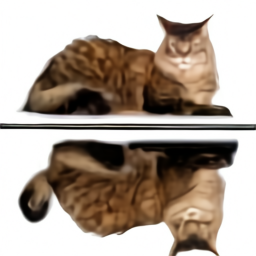}} 
 & \fbox{\includegraphics[scale=0.27]{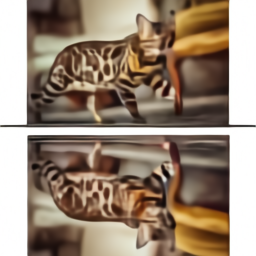}} 
 \\
\end{tabular}}%
\hspace{0.75mm}%
\subfloat[``2 panel image of the exact same cat. on the top, a photo of the cat. on the bottom, an extreme close-up view of the cat in the photo.'']{%
\setlength\extrarowheight{-3pt}
\begin{tabular}{@{}c@{\hskip 0.05in}c@{}}
  \fbox{\includegraphics[scale=0.27]{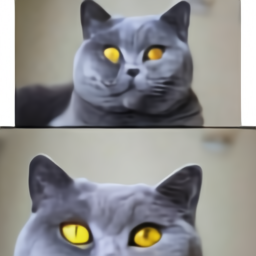}}
  & \fbox{\includegraphics[scale=0.27]{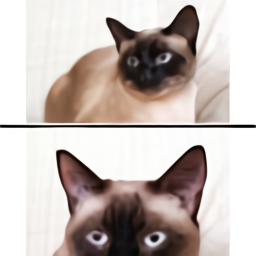}} 
 \\
  \fbox{\includegraphics[scale=0.27]{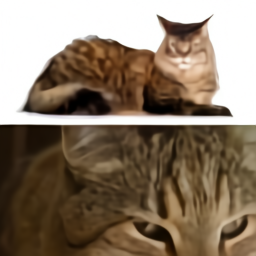}} 
 & \fbox{\includegraphics[scale=0.27]{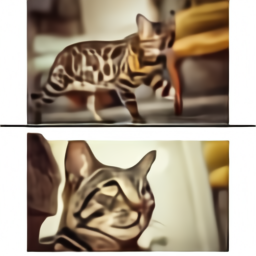}} 
 \\
\end{tabular}} \\
\subfloat[``the exact same cat on the top colored red on the bottom'']{%
\setlength\extrarowheight{-3pt}
\begin{tabular}{@{}c@{\hskip 0.05in}c@{}}
  \fbox{\includegraphics[scale=0.27]{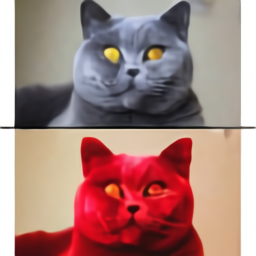}}
  & \fbox{\includegraphics[scale=0.27]{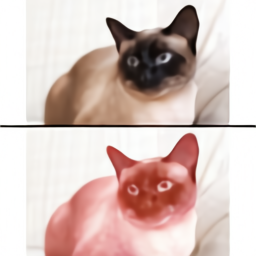}} 
 \\
  \fbox{\includegraphics[scale=0.27]{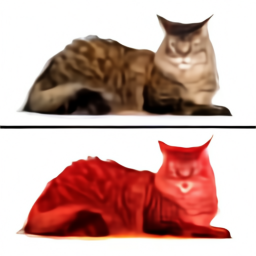}} 
 & \fbox{\includegraphics[scale=0.27]{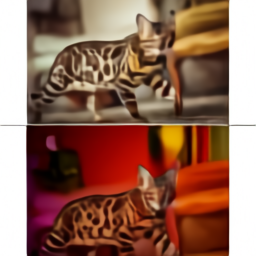}} 
 \\
\end{tabular}}%
\hspace{0.75mm}%
\subfloat[``2 panel image of the exact same cat. on the top, a photo of the cat. on the bottom, the cat with sunglasses.'']{%
\setlength\extrarowheight{-3pt}
\begin{tabular}{@{}c@{\hskip 0.05in}c@{}}
  \fbox{\includegraphics[scale=0.27]{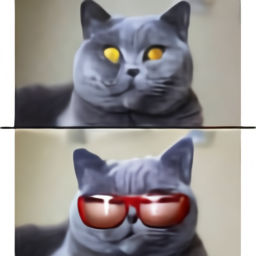}}
  & \fbox{\includegraphics[scale=0.27]{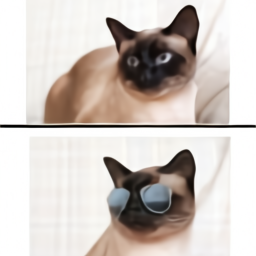}} 
 \\
  \fbox{\includegraphics[scale=0.27]{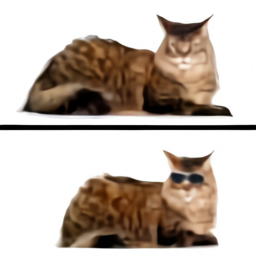}} 
 & \fbox{\includegraphics[scale=0.27]{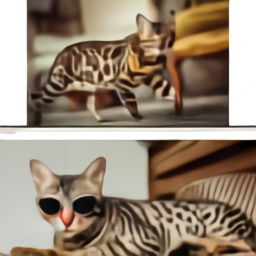}} 
 \\
\end{tabular}}%
\hspace{0.75mm}%
\subfloat[``the exact same cat on the top as a postage stamp on the bottom'']{%
\setlength\extrarowheight{-3pt}
\begin{tabular}{@{}c@{\hskip 0.05in}c@{}}
  \fbox{\includegraphics[scale=0.27]{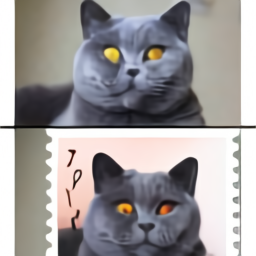}}
  & \fbox{\includegraphics[scale=0.27]{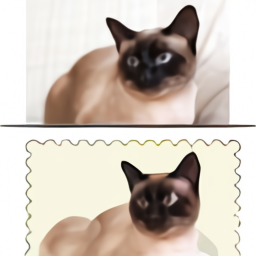}} 
 \\
  \fbox{\includegraphics[scale=0.27]{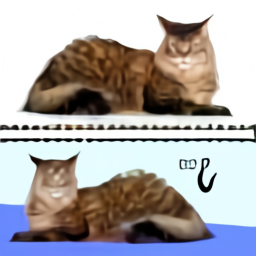}} 
 & \fbox{\includegraphics[scale=0.27]{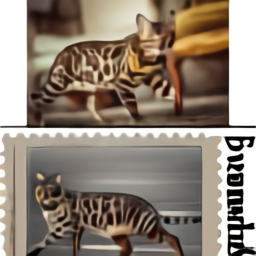}} 
 \\
\end{tabular}}%
\caption{Further examples of zero-shot image-to-image translation.}
\label{fig:zero_shot_samples}
\end{figure*}
Figure~\ref{fig:zero_shot_samples} shows further examples of zero-shot image-to-image translation, which we discussed in Section~\ref{sec:qual_findings}. We did not anticipate that this capability would emerge, and made no modifications to the training procedure to encourage it.

\end{document}